\documentclass[10pt,journal,compsoc]{IEEEtran}
\usepackage{algorithm}
\usepackage{algorithmic}
\usepackage{csquotes} 
\usepackage{color}
\usepackage{paralist}
\usepackage{latexsym}
\usepackage{indentfirst}
\usepackage{subfigure}
\usepackage{float}
\usepackage{cite}
\usepackage{CJKutf8}
\usepackage{multirow}
\usepackage{makecell}
\usepackage{mathrsfs}
\usepackage[colorlinks, linkcolor=red,  anchorcolor=blue, citecolor=blue]{hyperref}
\usepackage{textcomp,booktabs}
\usepackage{amssymb}
\usepackage{pifont}
\usepackage{epsfig}
\usepackage{colortbl}
\usepackage{amsmath} 
\usepackage{siunitx}
\usepackage{longtable}
\definecolor{mygray}{gray}{.9}
\definecolor{mypink}{rgb}{.99,.91,.95}
\definecolor{mycyan}{cmyk}{.3,0,0,0}
\hypersetup{%
        anchorcolor=blue, 
        colorlinks,
        breaklinks=true,
        plainpages=false,%
        citecolor=blue,
        linkcolor=red,
        urlcolor=magenta,
        bookmarksopen=true,%
        bookmarksnumbered=false,%
        bookmarksdepth=5%
}

\usepackage{booktabs} 
\usepackage{makecell} 
\usepackage{lipsum} 

\usepackage{colortbl}
\definecolor{mygray}{gray}{.9}
\definecolor{mypink}{rgb}{.99,.91,.95}
\definecolor{mycyan}{cmyk}{.3,0,0,0}

\usepackage{color}
\usepackage{xcolor}
\definecolor{citecolor}{HTML}{0071bc} 
\definecolor{SeaGreen4}{RGB}{0,205,102} 
\definecolor{SlateBlue}{RGB}{106,90,205} 
\definecolor{DarkRed}{RGB}{178,34,34} 
\usepackage{ragged2e}

\begin{document}
\begin{CJK}{UTF8}{gbsn}


\title{
    Leveraging Complementary Event Frame and Voxel Representations for RGB-Guided Multimodal Pre-training 
}

\author{ Wentao Wu, Xiao Wang*, \textit{Member, IEEE}, Yi Wang, 
    Chenglong Li*, \textit{Senior Member, IEEE}, \\ Yonghong Tian, \textit{Fellow, IEEE}, Bin Luo  

\IEEEcompsocitemizethanks{
\IEEEcompsocthanksitem Wentao Wu, Chenglong Li are with the School of Artificial Intelligence, Anhui University, Hefei, China (email: wa22201027@stu.ahu.edu.cn, lcl1314@foxmail.com) 
\IEEEcompsocthanksitem Xiao Wang, Yi Wang, and Bin Luo are with the School of Computer Science and Technology, Anhui University, Hefei, China. (email: \{xiaowang, luobin\}@ahu.edu.cn, yiwang7708@163.com) 
\IEEEcompsocthanksitem Yonghong Tian is with Peng Cheng Laboratory, Shenzhen, China, School of Computer Science, Peking University, China, and Shenzhen Graduate School, Peking University, China (email: yhtian@pku.edu.cn) 
\IEEEcompsocthanksitem Corresponding Author: Xiao Wang $\&$ Chenglong Li
}}

\markboth{IEEE Transactions on Pattern Analysis and Machine Intelligence}%
{Shell \MakeLowercase{\textit{et al.}}: Bare Demo of IEEEtran.cls for Computer Society Journals}

\IEEEtitleabstractindextext{%
\begin{abstract} 
\justifying
Existing Event-based multimodal pre-training methods primarily achieve joint multimodal learning by mapping Event features into the feature spaces of pre-trained models. However, these methods mainly focus on feature associations within existing representation spaces, lacking unified modeling of continuous RGB-Event observations and sufficient learning of cross-modal dependencies, thereby limiting the multimodal representational capacity of pre-trained models.
To address these issues, we propose CM3AE 2.0, a unified RGB Frame and Event-Voxel/-Frame pre-training framework. By jointly learning multimodal spatiotemporal representations and exploiting latent cross-modal dependencies through cross-modal generation, the framework enhances multimodal information interaction and collaborative modeling capabilities.
Specifically, we first propose an Event-aware Spatiotemporal Masked Reconstruction (ESMR) strategy to learn unified multimodal spatiotemporal representations. It leverages event responses to guide masking toward dynamically salient regions. Through spatiotemporal masked reconstruction, it learns spatial structures and their temporal variations from continuous RGB-Event observations. In addition, it incorporates multimodal fusion reconstruction and cross-modal contrastive learning to facilitate the modeling of cross-modal complementary information and the learning of shared representations.
Building upon this, we further propose a Hierarchical Cross-modal Generation (HCG) strategy to learn latent cross-modal dependencies through visual and language generation. At the visual level, Event-conditioned RGB generation is performed using a pre-trained diffusion model to capture visual associations between the two modalities. At the language level, video-level text description generation based on joint RGB-Event representations establishes connections between multimodal visual features and linguistic semantics, further enriching the learning of cross-modal dependencies.
Extensive experiments on four diverse downstream tasks demonstrate that CM3AE 2.0 consistently improves performance, validating the effectiveness and generalization capability of the proposed pre-training framework.
The source code of this paper is available at \url{https://github.com/Event-AHU/CM3AE}. 
\end{abstract}

\begin{IEEEkeywords}
Pre-trained Vision Model; Masked Auto-Encoder; Multimodal Learning; Vision and Language; Self-Supervised Learning
\end{IEEEkeywords}}

\maketitle

\IEEEdisplaynontitleabstractindextext

%
\IEEEpeerreviewmaketitle

\IEEEraisesectionheading{\section{Introduction}\label{sec::introduction}}
\IEEEPARstart{E}{vent} cameras, characterized by high temporal resolution, high dynamic range, low latency, and low power consumption, have become an important sensing modality for visual perception~\cite{gallego2020event}. Unlike conventional frame-based cameras, event cameras asynchronously capture pixel-level brightness changes, providing fine-grained motion cues and strong robustness to rapid motion and challenging illumination conditions. These properties have enabled a wide range of applications, including action recognition~\cite{wang2024hardvs}, object detection~\cite{zhang2025detecting}, visual object tracking~\cite{messikommer2025data}, semantic understanding~\cite{kong2024openess}, and image reconstruction~\cite{yang2025unsp}. However, most event-based perception models are still developed for individual tasks and rely heavily on task-specific annotations, motivating the development of transferable representations through large-scale pre-training.

\begin{figure*}[!htp]
\center
\includegraphics[width=0.95\textwidth]{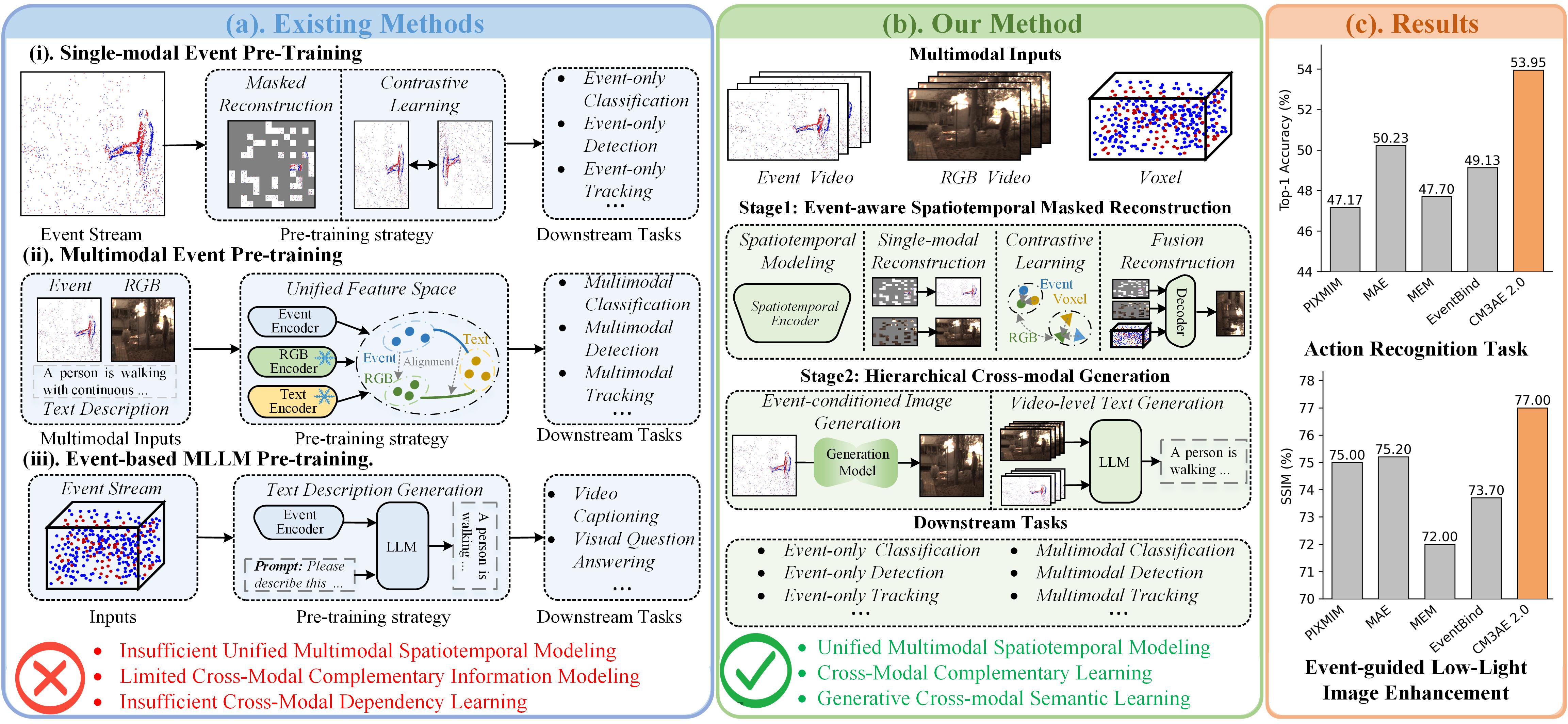}
\caption{Comparison of existing pre-training paradigms and CM3AE 2.0. (a) Representative Event-related pre-training paradigms. (b) Overview of the proposed CM3AE 2.0 framework. (c) Performance comparisons on action recognition and event-guided low-light image enhancement.} 
\label{fig:firstimg}
\end{figure*}

Large-scale self-supervised pre-training has demonstrated strong transferability across conventional vision tasks~\cite{wang2023MMPTMs, wang2026state}. Recent vision foundation models, including MAE~\cite{he2022mae}, iBOT~\cite{zhouimage}, DINOv2~\cite{oquab2023dinov2}, and BEiT v2~\cite{peng2022beit}, have shown the effectiveness of learning generalizable visual representations from large-scale unlabeled data. Extending to sequential visual understanding, VideoMAE~\cite{tong2022videomae} and VideoMAE V2~\cite{wang2023videomae} further explore temporal representation learning. Meanwhile, multimodal pre-training methods, such as CLIP~\cite{radford2021learning} and MultiMAE~\cite{bachmann2022multimae}, exploit cross-modal information to enhance representation learning. 
Inspired by these advances, increasing efforts have been devoted to event-based pre-training. 
Early studies primarily focus on Event representation learning through masked reconstruction~\cite{klenk2024masked} and contrastive learning~\cite{yang2023event}. Subsequent studies further explore spatiotemporal representation learning. Huang \textit{et al.}~\cite{huang2024data} introduce disentangled masked modeling to learn local spatiotemporal features and global representations, while TESPEC~\cite{mohammadi2025tespec} investigates long-term temporal modeling through pseudo-grayscale video reconstruction. Recently, researchers have increasingly leveraged pre-trained foundation models for cross-modal learning. EventBind~\cite{zhou2024eventbind} aligns Event representations with the shared image-text space of CLIP, while Jeong \textit{et al.}~\cite{jeong2026cross} and ScaleEvent~\cite{chen2026scaling} transfer visual knowledge through cross-modal alignment and distillation. Furthermore, EventGPT~\cite{liu2025eventgpt} incorporates Event representations into multimodal large language models for semantic understanding.

Despite these advances, three key challenges remain in unified RGB-Event pre-training, as illustrated in Fig.~\ref{fig:firstimg}. \textbf{First,} although existing studies have explored Event spatiotemporal learning~\cite{huang2024data,mohammadi2025tespec} and cross-modal representation alignment~\cite{zhou2024eventbind}, unified modeling of temporal dynamics across continuous RGB-Event observations remains insufficient. Existing Event-based methods have explored temporal modeling within Event streams, while cross-modal alignment methods primarily focus on establishing correspondences between Event representations and pre-trained visual or vision-language features. Consequently, the temporal evolution of RGB appearance and Event activity is not fully exploited within a unified multimodal pre-training framework.
\textbf{Second,} existing approaches such as EventBind~\cite{zhou2024eventbind} and ScaleEvent~\cite{chen2026scaling} primarily transfer knowledge from pre-trained models through feature alignment and distillation, with limited explicit integration of complementary information from paired RGB and Event observations. Consequently, complementary cues from RGB appearance and Event motion may not be sufficiently combined, while shared features across modalities may not be adequately learned.
\textbf{Third,} although existing studies have begun to investigate semantic understanding of Event data~\cite{liu2025eventgpt}, the latent dependencies among Event dynamics, RGB visual content, and scene semantics remain insufficiently modeled, limiting the ability to exploit information that can be inferred across modalities.

To address these limitations, we propose CM3AE 2.0, a unified RGB Frame and Event-Voxel/-Frame pre-training framework. The framework learns unified multimodal spatiotemporal representations and further explores latent cross-modal dependencies through cross-modal generation, enhancing multimodal information interaction and collaborative modeling.
Specifically, to address the insufficient modeling of temporal dynamics in continuous RGB-Event observations and the limited exploitation of complementary information, we introduce an Event-aware Spatiotemporal Masked Reconstruction (ESMR) strategy. It leverages event responses to guide masking toward dynamically salient regions, and employs modality-specific spatiotemporal masked reconstruction to learn spatial structures and their temporal variations from continuous RGB-Event observations. To further preserve Event-specific information, the event-specific reconstruction loss supervises event occurrence and polarity prediction. Meanwhile, multimodal fusion reconstruction integrates RGB, Event, and Voxel information to exploit complementary cues, while cross-modal contrastive learning promotes shared representation learning across modalities.
To further explore latent dependencies among Event dynamics, RGB visual content, and scene semantics, we introduce a Hierarchical Cross-modal Generation (HCG) strategy. At the visual level, event-conditioned RGB image generation uses Event representations to condition a pre-trained diffusion model, transferring visual generative priors to Event representations and encouraging them to capture RGB appearance and structural information. At the language level, video-level text generation leverages joint RGB-Event representations to guide a pre-trained language model in generating scene descriptions, establishing connections between multimodal visual features and linguistic semantics.

To comprehensively evaluate the transferability and generalization capability of the proposed framework, we conduct extensive experiments under both Event-only and RGB-Event multimodal settings, covering four representative downstream tasks, including action recognition, object detection, visual object tracking, and event-guided low-light image enhancement. Experimental results demonstrate that CM3AE 2.0 provides effective pre-trained initialization for diverse downstream architectures and consistently improves the representation capability of both Event and RGB-Event models.

To sum up, the contributions of this work can be summarized as follows:
\begin{itemize}
    \item We propose CM3AE 2.0, a unified RGB Frame and Event-Voxel/-Frame pre-training framework that learns unified multimodal spatiotemporal representations and further explores latent cross-modal dependencies through cross-modal generation.

    \item We propose an Event-aware Spatiotemporal Masked Reconstruction strategy that learns unified multimodal spatiotemporal representations by modeling temporal dynamics in continuous RGB-Event observations and exploiting complementary information and shared features through fusion reconstruction and cross-modal contrastive learning.

    \item We propose a Hierarchical Cross-modal Generation strategy that combines event-conditioned RGB image generation and video-level text generation to explore latent cross-modal dependencies by leveraging visual generative priors and linguistic knowledge.

    \item We conduct extensive experiments on four downstream tasks, demonstrating the effectiveness of CM3AE 2.0 across diverse perception tasks, as well as the strong transferability and generalization capability of the learned representations.
\end{itemize} 

A preliminary version of this work appeared in the proceedings of ACM Multimedia 2025 (ACM MM 2025) under the title CM3AE~\cite{wu2025cm3ae}. Compared with the conference version, this journal extension substantially advances the original work in three aspects. 
\textbf{(1)} We extend the original frame-level multimodal pre-training framework to support unified multimodal spatiotemporal representation learning. Specifically, we introduce an Event-aware spatiotemporal masked reconstruction strategy. Building upon the original framework, this strategy leverages Event information to guide masking toward dynamically salient regions, employs spatiotemporal masked reconstruction to model spatiotemporal relationships across continuous RGB-Event observations, and incorporates an event-specific reconstruction loss to learn Event-specific structural information, thereby enabling unified multimodal spatiotemporal representation learning. 
\textbf{(2)} We introduce a new hierarchical cross-modal generation strategy, which combines Event-conditioned RGB image generation using a pre-trained diffusion model with video-level text generation using a pre-trained language model. By leveraging visual generative priors and linguistic knowledge, this strategy explores latent cross-modal dependencies among Event dynamics, RGB appearance, and high-level scene semantics. 
\textbf{(3)} We extend the evaluation scope by introducing an event-guided low-light image enhancement task and conducting experiments across diverse datasets under both RGB-Event and Event-only settings. Furthermore, comprehensive ablation studies, transferability analyses, computational efficiency evaluations, and qualitative visualizations validate the effectiveness and generalization capability of the proposed framework.
Overall, CM3AE 2.0 goes beyond the original frame-level cross-modal learning framework by establishing a two-stage pre-training paradigm that integrates unified multimodal spatiotemporal representation learning with generative supervision for learning latent cross-modal dependencies.

\section{Related Work}\label{sec::relatedwork}

\subsection{Event-based Vision}  
\label{subsec:event_based_vision}

Research on event-based visual perception primarily focuses on modeling the spatial structure and temporal dynamics of event streams, with applications in recognition, detection, and tracking. For recognition, EventMamba~\cite{ren2025rethinking} learns event representations through hierarchical point-based processing and state-space modeling, while EMP~\cite{cao2025exploiting} combines accumulated and reconstructed event frames for action recognition. Wang \emph{et al.}~\cite{wang2024hardvs,wang2026event} introduce the HARDVS and CeleX-HAR benchmarks and propose ESTF and EVMamba, respectively, to learn spatiotemporal features. For object detection, RVTs~\cite{gehrig2023recurrent} and SATE~\cite{zhu2026rethinking} combine spatial feature extraction with recurrent temporal modeling, whereas MvHeat-DET~\cite{wang2025object} explores event feature extraction using mixture-of-experts heat-conduction operators. For tracking and segmentation, Wang \emph{et al.}~\cite{wang2024event} introduce the EventVOT benchmark and HDETrack for event-only tracking, while Zhu \emph{et al.}~\cite{zhu2024continuous} predict subsequent object masks using recurrent temporal embeddings and historical event-mask references.
Event streams have also been used to assist conventional image enhancement, improving image quality under challenging illumination conditions. EvLight~\cite{liang2024towards} performs low-light image enhancement through signal-to-noise ratio-guided regional feature selection, while Weng \emph{et al.}~\cite{weng2024event} address local overexposure and underexposure in high-dynamic-range scenes. More recently, RetinEV~\cite{sun2025low} estimates illumination from event timestamps to improve low-light image enhancement. These studies demonstrate the potential of event-based vision across diverse perception tasks and highlight the importance of effective spatial and temporal feature modeling for improving event-based visual perception.

\subsection{RGB-Event based Fusion}  
\label{subsec:rgb_event_fusion}

RGB frames provide rich appearance and semantic information but are vulnerable to low illumination and motion blur, whereas event streams capture rapid motion with high temporal resolution but offer limited appearance information. Their complementarity motivates RGB-Event fusion, although modality discrepancies and spatiotemporal misalignment remain challenging.
For object tracking, MAMTrack~\cite{sun2025exploring} combines Mamba-based fusion with historical information to model target appearance changes. AlignTrack~\cite{sun2026aligntrack} leverages target-related priors to guide cross-modal feature alignment, while CRSOT~\cite{zhu2025crsot} employs uncertainty-aware fusion to address spatiotemporal misalignment between sensors. For object detection, FlexEvent~\cite{lu2026flexevent} adaptively fuses RGB semantics with high-frequency event observations to support detection at varying operational frequencies. Beyond Duality~\cite{wang2026beyond} explicitly disentangles shared and modality-specific features, enabling adaptive exploitation of RGB appearance details and event-based motion cues under different scene conditions. For recognition, MMHCO-HAR~\cite{wang2026human} combines heat-conduction-based feature interaction with policy-driven adaptive fusion for human activity recognition. Wang \emph{et al.}~\cite{wang2026rgb} introduce the EventPAR benchmark and an asymmetric RWKV fusion framework, extending RGB-Event fusion to pedestrian attribute recognition, including appearance and emotion attributes. 
For optical flow estimation, Diff-ABFlow~\cite{wang2026injecting} leverages appearance cues from RGB frames and boundary information from event streams to condition diffusion-based estimation.
Overall, existing methods primarily address modality discrepancies within downstream tasks. However, the limited scale and diversity of downstream datasets constrain the generalizability of the learned representations.



\begin{figure*}
\centering
\includegraphics[width=0.95\textwidth]{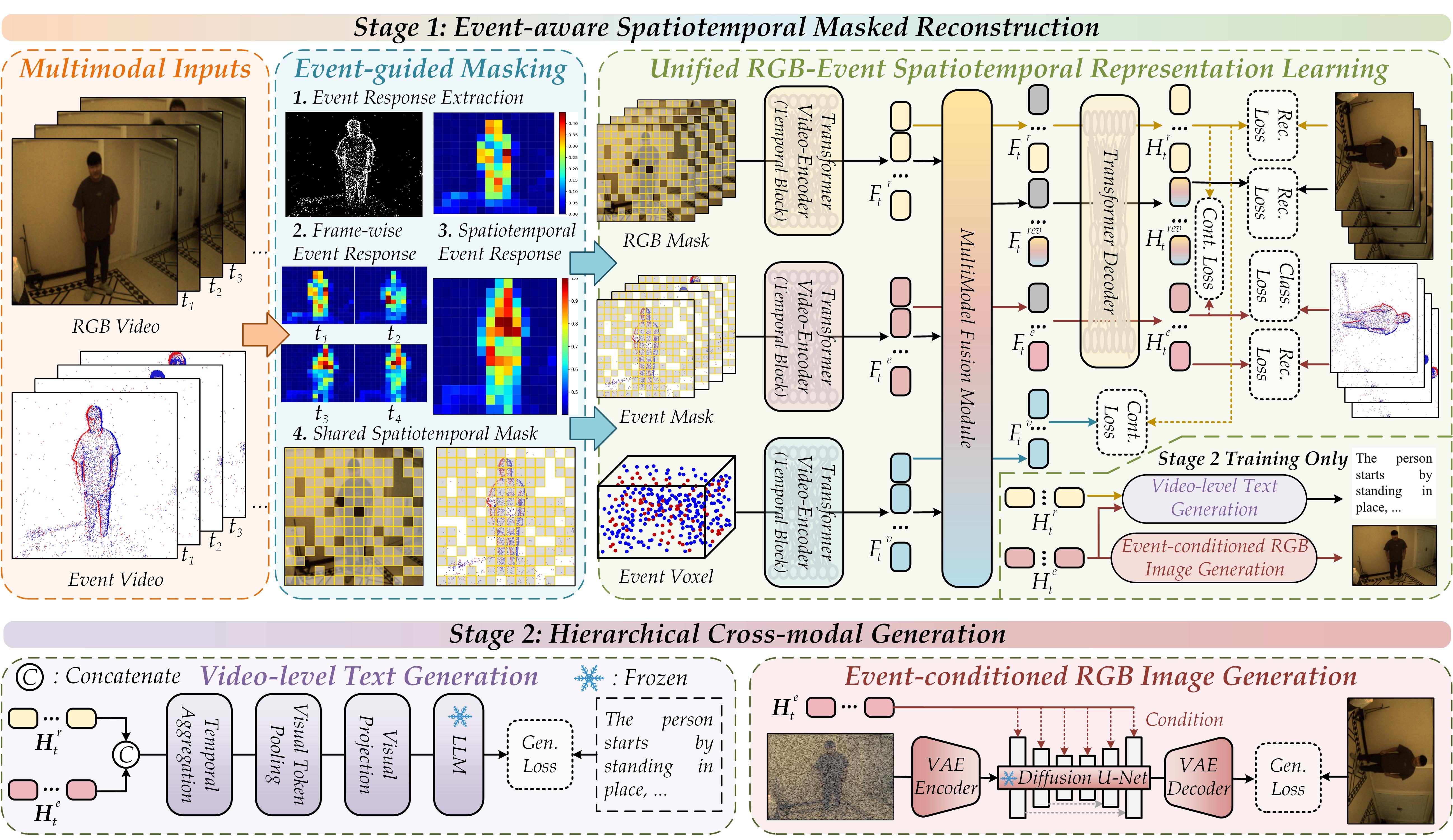}
\caption{Overview of the proposed CM3AE 2.0 framework, comprising two stages: (1) Event-aware Spatiotemporal Masked Reconstruction (ESMR), which learns unified multimodal spatiotemporal representations by modeling cross-frame dynamics and exploiting complementary information and shared features across modalities; and (2) Hierarchical Cross-modal Generation (HCG), which explores latent cross-modal dependencies through Event-conditioned RGB Image Generation and Video-level Text Generation.}
\label{fig:framework} 
\end{figure*}

\subsection{Pre-trained Foundation Models}  
\label{subsec:pretrained_foundation_models}

Early pre-training primarily focused on RGB images or videos. For example, MoCo~\cite{he2020momentum} learns discriminative visual representations through contrastive learning, MAE~\cite{he2022mae} learns image representations via pixel reconstruction, and VideoMAE V2~\cite{wang2023videomae} extends masked modeling to video pre-training. With the development of multimodal pre-training, CLIP~\cite{radford2021learning} achieves cross-modal alignment through image-text contrastive learning, MultiMAE~\cite{bachmann2022multimae} learns cross-modal representations by jointly reconstructing RGB, depth, and semantic information, while VehicleMAE~\cite{wu2026vehicle} leverages vehicle structural priors and image-text knowledge to guide vehicle appearance reconstruction. Recent studies further explore autoregressive modeling and generative priors. AIMv2~\cite{fini2025multimodal} learns visual representations by autoregressively predicting image patches and text tokens, while CleanDIFT~\cite{stracke2025cleandift} and VLV~\cite{zhang2026vision} enhance visual representations through diffusion feature distillation and generative model knowledge transfer.

Inspired by these studies, researchers have begun to explore pre-training methods for event data. Early works primarily focused on event representation learning. ECDP~\cite{yang2023event} constructs contrastive learning objectives using paired images, MEM~\cite{klenk2024masked} learns event representations through masked event modeling, and TESPEC~\cite{mohammadi2025tespec} captures temporal information using long event sequences and pseudo-grayscale video reconstruction.
Subsequent studies have expanded toward multimodal learning. Liang \emph{et al.}~\cite{liang2025efficient} transfer pre-trained image knowledge to the event domain through adaptive prompting, ScaleEvent~\cite{chen2026scaling} employs structure-aware distillation from visual foundation models, and EventBind~\cite{zhou2024eventbind} aligns event representations with pre-trained image-text feature spaces. CM3AE~\cite{wu2025cm3ae} explores joint RGB-Event representation learning through multimodal reconstruction and cross-modal contrastive learning. Furthermore, EventGPT~\cite{liu2025eventgpt} and EventDrive~\cite{lu2026eventdrive} investigate the use of language knowledge for event-stream understanding and the integration of RGB, events, and language for driving perception and reasoning, respectively. However, existing studies primarily focus on pre-trained knowledge transfer and cross-modal semantic associations, leaving the unified spatiotemporal modeling of continuous RGB-Event observations insufficiently explored.

\section{Methodology} \label{sec3::Methodology}

\subsection{Overview} 

As illustrated in Fig.~\ref{fig:framework}, we propose CM3AE 2.0, a unified RGB Frame and Event-Voxel/-Frame pre-training framework. The framework consists of two stages, Event-aware Spatiotemporal Masked Reconstruction (ESMR) and Hierarchical Cross-modal Generation (HCG), aiming to learn unified multimodal spatiotemporal representations and explore latent cross-modal dependencies.
In the first stage, to effectively model spatiotemporal relationships and exploit cross-modal complementary information from continuous RGB-Event observations, we introduce ESMR. Specifically, event-guided masking identifies dynamically salient regions based on Event activity, while spatiotemporal masked reconstruction captures spatiotemporal relationships through spatial encoding and cross-frame interactions. An event-specific reconstruction loss further enhances the representation of Event-specific information. Moreover, multimodal fusion reconstruction exploits complementary information by integrating RGB, Event, and Voxel features, while cross-modal contrastive learning promotes shared representation learning through feature alignment.
In the second stage, to further explore cross-modal associations and high-level semantic knowledge that are insufficiently captured by feature-level learning, we introduce HCG. Specifically, event-conditioned RGB image generation leverages visual generative priors from a pre-trained diffusion model to establish associations between Event dynamics and RGB appearance. Meanwhile, video-level text generation generates video descriptions from joint RGB-Event representations, establishing connections between multimodal visual features and high-level linguistic semantics.

\subsection{CM3AE 2.0 Pre-training Framework}   

The first stage consists of four components, i.e., event-guided masking, spatiotemporal masked reconstruction, multimodal fusion reconstruction, and cross-modal contrastive learning.

\noindent $\bullet$ \textbf{Stage 1: Event-aware Spatiotemporal Masked Reconstruction.}
Existing RGB-Event pre-training methods lack unified modeling of spatiotemporal relationships across continuous RGB-Event observations and fail to fully exploit complementary information between modalities. To address these limitations, we develop an Event-aware Spatiotemporal Masked Reconstruction (ESMR) strategy, which consists of four components: event-guided masking, spatiotemporal masked reconstruction, multimodal fusion reconstruction, and cross-modal contrastive learning. The overall objective of Stage 1 is to learn unified multimodal spatiotemporal representations. Specifically, event-guided masking and spatiotemporal masked reconstruction model temporal dynamics and cross-frame relationships in continuous RGB-Event observations. Meanwhile, multimodal fusion reconstruction exploits complementary information across modalities, while cross-modal contrastive learning promotes shared representation learning.

\textbf{\textit{Event-guided Masking.}}
Given a paired RGB-Event sequence of $T$ frames, we denote the RGB and Event inputs as
$\mathcal{I}^{r}=\{I_t^{r}\}_{t=1}^{T}$ and
$\mathcal{I}^{e}=\{I_t^{e}\}_{t=1}^{T}$, respectively.
Each frame is divided into $N=196$ non-overlapping patches and projected into $D=768$-dimensional patch embeddings using modality-specific patch embedding layers.

Unlike conventional random masking that assigns equal masking probabilities to all spatial patches, our strategy exploits Event activity to identify dynamically informative regions.
Specifically, positive and negative events are first converted into a binary Event activity map, where pixels with Event activity are assigned a value of one and inactive pixels zero.
Based on this map, the frame-wise Event response $s_{t,i}$ is defined as the proportion of active Event pixels within the $i$-th patch of the $t$-th frame, quantifying local Event activity at each spatial position.

For genuine video sequences, we aggregate frame-wise responses across valid frames to capture both persistent motion and short-term salient variations:
\begin{equation}
s_i =
\alpha\,\operatorname{Mean}_{t\in\mathcal{V}}(s_{t,i})
+
(1-\alpha)\,\operatorname{Max}_{t\in\mathcal{V}}(s_{t,i}),
\end{equation}
where $\mathcal{V}$ denotes the set of valid frames and $\alpha=0.7$.
The resulting response $s_i$ is normalized across spatial positions to obtain $\bar{s}_i$, reflecting the relative Event activity at each position.

Based on these responses, we perform event-guided probabilistic masking.
The sampling weight of the $i$-th patch is defined as:
\begin{equation}
w_i =
(1-\gamma)
+
\gamma\bar{s}_i
+
0.01\eta\xi_i,
\end{equation}
where $\gamma$ controls the strength of Event guidance, $\xi_i\sim\mathcal{U}(0,1)$ introduces a small stochastic perturbation, and $\eta$ controls its magnitude. We set $\gamma=0.6$ and $\eta=0.1$. 
Patches with stronger Event responses are assigned larger sampling weights and are therefore more likely to be selected as masked reconstruction targets. The uniform baseline term prevents the sampling weights from being determined solely by Event activity, while the stochastic perturbation introduces small random variations. The default masking ratio is set to $75\%$, leaving the remaining $25\%$ of patches visible.

To preserve spatial correspondence, paired RGB and Event inputs share the same masking pattern. For genuine video sequences, temporally aggregated Event responses guide a shared spatial mask across frames, whereas each image in an image-level pseudo-sequence is independently masked based on its own Event response.

\textbf{\textit{Spatiotemporal Masked Reconstruction.}}
After event-guided masking, the visible patch embeddings of RGB and Event are obtained as
$\mathbf{E}_t^r,\mathbf{E}_t^e\in\mathbb{R}^{49\times768}$, respectively.
We add spatial and temporal positional embeddings to these features and prepend a CLS token, yielding the encoder inputs
$\widetilde{\mathbf{E}}_t^r,\widetilde{\mathbf{E}}_t^e\in\mathbb{R}^{50\times768}$.
They are then fed into two modality-specific ViT encoders, each consisting of 12 Transformer blocks.
Temporal attention is inserted after every four blocks to model spatiotemporal relationships across consecutive frames.
Specifically, for modality $m\in\{r,e\}$, the features at the same spatial position across $T$ consecutive frames are organized into
$\mathbf{Q}_i^m\in\mathbb{R}^{T\times768}$.
Cross-frame temporal interaction is formulated as
\begin{equation}
\widetilde{\mathbf{Q}}_i^m
=
\mathbf{Q}_i^m
+
\operatorname{TAttn}
\left(
\operatorname{LN}(\mathbf{Q}_i^m)
\right),
\end{equation}
where $\operatorname{TAttn}$ denotes temporal self-attention.
After spatial encoding and cross-frame temporal interaction, the RGB and Event encoder outputs are obtained as $\mathbf{F}_t^r$ and $\mathbf{F}_t^e$, respectively.
For image-level pseudo-sequences without genuine temporal continuity, cross-frame temporal attention is disabled.

We further introduce a Voxel branch.
Given a Voxel sequence $\mathcal{X}^{v}=\{\mathbf{V}_t\}_{t=1}^{T}$, the raw Voxel features of each frame are divided into 196 patches and projected into a 768-dimensional feature space, yielding the patch embeddings
$\mathbf{E}_t^v\in\mathbb{R}^{196\times768}$.
Unlike the RGB and Event branches, the Voxel branch is not spatially masked, and all patches are retained.
After adding a CLS token and spatial and temporal positional embeddings, the resulting features
$\widetilde{\mathbf{E}}_t^v\in\mathbb{R}^{197\times768}$
are fed into a separate ViT encoder.
Spatial encoding and temporal interaction produce the complete Voxel representation $\mathbf{F}_t^v$.
Cross-frame temporal interaction is applied only to valid video frames with Voxel observations.
The RGB and Event encoder outputs, $\mathbf{F}_t^r$ and $\mathbf{F}_t^e$, are then projected from 768 to 512 dimensions through separate linear projection layers.
We insert 147 learnable mask tokens at the original masked positions, yielding the decoder inputs
$\widehat{\mathbf{E}}_t^r,\widehat{\mathbf{E}}_t^e\in\mathbb{R}^{197\times512}$.
These inputs are fed into the corresponding RGB and Event decoders, each consisting of eight Transformer blocks.
The decoders produce the feature representations $\mathbf{H}_t^r$ and $\mathbf{H}_t^e$, from which linear prediction layers generate the reconstructed RGB and Event contents, denoted by $\hat{\mathbf{X}}_t^r$ and $\hat{\mathbf{X}}_t^e$, respectively.
Reconstruction losses are computed only over masked patches.

For the Event branch, we further introduce an event-specific reconstruction loss.
In addition to pixel-level Event reconstruction, two lightweight prediction heads are applied to predict event occurrence and dominant polarity in masked patches.
The occurrence head predicts whether a masked patch contains valid Events, whereas the polarity head predicts the dominant polarity only for patches containing valid Events.
These objectives enable the model to capture Event-specific structural information beyond pixel-level reconstruction.

\textbf{\textit{Multimodal Fusion Reconstruction.}}
To further exploit complementary information across modalities, we perform multimodal fusion reconstruction using the RGB, Event, and Voxel encoder features, $\mathbf{F}_t^r$, $\mathbf{F}_t^e$, and $\mathbf{F}_t^v$.
Specifically, the visible RGB and Event features are concatenated and fed into a Transformer-based fusion module, where multimodal self-attention is used to obtain the RGB-Event fused representation $\mathbf{F}_t^{re}$.
For RGB-Event-Voxel fusion, the same fusion process is adopted by combining the visible RGB and Event features with the complete Voxel patch features, yielding the joint representation $\mathbf{F}_t^{rev}$.
The two fused representations are then separately projected to 512 dimensions.
By inserting 147 learnable mask tokens at the original masked positions, we obtain the decoder inputs
$\widehat{\mathbf{E}}_t^{re},\widehat{\mathbf{E}}_t^{rev}\in\mathbb{R}^{197\times512}$.
Both branches use the same RGB decoder to generate the reconstructed RGB contents $\hat{\mathbf{X}}_t^{re}$ and $\hat{\mathbf{X}}_t^{rev}$, respectively.
The reconstruction losses are computed over masked RGB patches to encourage the model to exploit complementary information across modalities.

\textbf{\textit{Cross-modal Contrastive Learning.}}
In addition to reconstruction, we employ cross-modal contrastive learning to promote shared representation learning across modalities.
Specifically, the RGB and Event decoder features, $\mathbf{H}_t^r$ and $\mathbf{H}_t^e$, are aggregated through learnable token pooling to obtain the frame-level representations
$\mathbf{z}_t^r,\mathbf{z}_t^e\in\mathbb{R}^{512}$.
After $L_2$ normalization, the similarity between RGB and Event representations is computed to construct a bidirectional contrastive objective.
Corresponding RGB-Event observations are treated as positive pairs, while representations from other observations serve as negatives.
To reduce false-negative interference, other frames from the same genuine video are excluded from the negative set, whereas different images in image-level pseudo-sequences are treated as independent observations.
Furthermore, frames with extremely low Event activity are filtered during RGB-Event contrastive learning.

Similarly, we construct an RGB-Voxel contrastive objective.
The complete Voxel encoder features $\mathbf{F}_t^v$ are aggregated through learnable token pooling and projected from 768 to 512 dimensions, yielding the Voxel representation $\mathbf{z}_t^v\in\mathbb{R}^{512}$.
The same contrastive learning procedure is then applied to RGB-Voxel pairs, using only frames with valid Voxel observations.
Through these components, Stage 1 learns unified multimodal spatiotemporal representations from continuous RGB-Event observations while exploiting complementary information and shared features across modalities.

\noindent $\bullet$ \textbf{Stage 2: Hierarchical Cross-modal Generation.}
Stage 1 learns unified multimodal spatiotemporal representations through reconstruction and contrastive learning.
To further explore cross-modal dependencies at the visual and semantic levels, we introduce a Hierarchical Cross-modal Generation (HCG) strategy consisting of event-conditioned RGB image generation and video-level text generation.
The former transfers visual generative priors from a pre-trained diffusion model to Event representations, while the latter introduces linguistic knowledge through video-level description generation.

\textbf{\textit{Event-conditioned RGB Image Generation.}}
To establish the relationship between Event representations and corresponding RGB appearance, we employ the pre-trained Stable Diffusion v1.4 model~\cite{rombach2022high}, conditioned on the Event decoder features $\mathbf{H}_t^e$.
After removing the CLS token, the remaining Event patch features are projected from 512 dimensions into the diffusion conditioning space through a learnable linear projection layer, yielding $\mathbf{C}_t^e\in\mathbb{R}^{196\times768}$.

For the corresponding RGB image $I_t^r$, a frozen VAE encoder extracts its latent representation $\mathbf{z}_t$.
Gaussian noise $\boldsymbol{\epsilon}_t\sim\mathcal{N}(0,\mathbf{I})$ is added at a randomly sampled timestep $\tau$ to obtain the noisy latent $\mathbf{z}_{t,\tau}$.
The frozen diffusion U-Net then predicts the injected noise conditioned on $\mathbf{C}_t^e$:
\begin{equation}
\begin{aligned}
\mathbf{z}_{t,\tau}
&=
\sqrt{\bar{\alpha}_{\tau}}\mathbf{z}_t
+
\sqrt{1-\bar{\alpha}_{\tau}}\boldsymbol{\epsilon}_t,\\
\hat{\boldsymbol{\epsilon}}_t
&=
\epsilon_\theta
\left(
\mathbf{z}_{t,\tau},
\tau,
\mathbf{C}_t^e
\right).
\end{aligned}
\end{equation}

During training, the VAE and diffusion U-Net remain frozen, while the noise-prediction objective updates the conditioning projection and upstream Event representations.
Through this process, the Event representations learn dependencies between Event dynamics and RGB appearance under the guidance of pre-trained visual generative priors.

\textbf{\textit{Video-level Text Generation.}}
To introduce high-level linguistic semantics into multimodal representation learning, we employ the RGB-Event fusion decoder features $\mathbf{H}_t^{re}$ and Event decoder features $\mathbf{H}_t^e$ obtained in Stage 1.
After removing the CLS tokens, the patch features from valid frames are concatenated and aggregated into eight visual tokens through adaptive token pooling.
A learnable projection layer further maps the aggregated features from 512 dimensions into the embedding space of a pre-trained language model, yielding $\mathbf{V}\in\mathbb{R}^{8\times4096}$:
\begin{equation}
\mathbf{V}
=
\psi_l
\left(
\operatorname{Pool}
\left(
\operatorname{Cat}_{t}
\left[
\mathbf{H}_{t,1:N}^{re};
\mathbf{H}_{t,1:N}^{e}
\right]
\right)
\right),
\end{equation}
where $\operatorname{Pool}(\cdot)$ denotes visual token aggregation and $\psi_l(\cdot)$ represents the visual-to-language projection.
The resulting visual tokens are combined with a textual prompt and fed into a pre-trained Llama-3-8B model to generate video-level descriptions.

Since the pre-training datasets lack textual annotations, we use Qwen2.5-VL~\cite{bai2025qwen25vl} to generate video-level descriptions offline as pseudo-text labels.
During training, the pre-trained Llama-3-8B model~\cite{grattafiori2024llama} remains frozen, and the autoregressive language modeling loss is computed only over the target description tokens.
The gradients are propagated through the visual-to-language projection to the upstream multimodal representations, enabling them to learn scene-level semantic relationships from linguistic supervision.

Through these two generation tasks, HCG models cross-modal dependencies at both the visual appearance and linguistic semantic levels.

\subsection{Training Objectives}
To learn unified multimodal spatiotemporal representations and further explore latent cross-modal dependencies, we formulate a two-stage training objective.
Stage 1 combines masked reconstruction and event-specific reconstruction loss to model temporal dynamics and strengthen Event-specific representations, while multimodal fusion reconstruction and cross-modal contrastive learning exploit complementary information and promote shared representation learning.
Stage 2 retains the reconstruction objectives and introduces visual and language generation to further explore latent cross-modal dependencies.

For masked reconstruction, let $\mathcal{M}_{t}$ denote the set of masked patches in the $t$-th frame, and let $\mathbf{X}_{t,i}^{r}$ and $\mathbf{X}_{t,i}^{e}$ denote the ground-truth RGB and Event patch contents, respectively. Based on the reconstruction predictions $\hat{\mathbf{X}}_{t}^{r}$ and $\hat{\mathbf{X}}_{t}^{e}$ defined above, the joint masked reconstruction loss is computed over the masked regions as:
\begin{equation}
\begin{aligned}
\mathcal{L}_{m}
=
\frac{1}{|\mathcal{M}|}
\sum_{t}\sum_{i\in\mathcal{M}_{t}}
\Big(
&\left\|
\mathbf{X}_{t,i}^{r}
-
\hat{\mathbf{X}}_{t,i}^{r}
\right\|_{2}^{2}
\\
+&
\left\|
\mathbf{X}_{t,i}^{e}
-
\hat{\mathbf{X}}_{t,i}^{e}
\right\|_{2}^{2}
\Big),
\end{aligned}
\end{equation}
where $|\mathcal{M}|$ denotes the total number of valid masked patches.

To explicitly preserve event-specific properties beyond pixel reconstruction, we introduce event occurrence and polarity prediction based on the Event decoder features. The event occurrence target $y_{t,i}^{\mathrm{occ}}$ indicates whether the $i$-th patch in the $t$-th frame contains any positive or negative event pixels. The polarity target $y_{t,i}^{\mathrm{pol}}$ is determined by the dominant event polarity, with a label of 1 assigned when the number of positive event pixels is greater than or equal to that of negative event pixels, and 0 otherwise.

Let $a_{t,i}$ denote the predicted event occurrence logit and $\mathbf{b}_{t,i}$ denote the logits for the two polarity classes. The event occurrence loss is computed using binary cross-entropy over masked patches in valid frames:
\begin{equation}
\begin{aligned}
\mathcal{L}_{\mathrm{occ}}
=
-\frac{1}{N_{\mathrm{occ}}}
\sum_{t\in\mathcal{V}}
\sum_{i\in\mathcal{M}_t}
\Big[
&y_{t,i}^{\mathrm{occ}}
\log \sigma(a_{t,i})\\
+&(1-y_{t,i}^{\mathrm{occ}})
\log\big(1-\sigma(a_{t,i})\big)
\Big],
\end{aligned}
\end{equation}
where $\sigma(\cdot)$ denotes the sigmoid function, $\mathcal{V}$ denotes the set of valid frames, $\mathcal{M}_t$ denotes the set of masked patches in the $t$-th frame, and $N_{\mathrm{occ}}$ denotes the total number of valid masked patches.

The polarity loss is computed using cross-entropy only for masked patches containing valid events:
\begin{equation}
\mathcal{L}_{\mathrm{pol}}
=
-\frac{1}{N_{\mathrm{pol}}}
\sum_{t\in\mathcal{V}}
\sum_{i\in\mathcal{M}_t}
y_{t,i}^{\mathrm{occ}}
\log
\operatorname{Softmax}(\mathbf{b}_{t,i})_{y_{t,i}^{\mathrm{pol}}},
\end{equation}
where $N_{\mathrm{pol}}$ denotes the number of masked patches containing valid events. The resulting event-specific reconstruction loss is defined as:
\begin{equation}
\mathcal{L}_{\mathrm{esp}}
=
\mathcal{L}_{\mathrm{occ}}
+
\mathcal{L}_{\mathrm{pol}}.
\end{equation}

For cross-modal complementary representation learning, the fused representation $\mathbf{F}_{t}^{re}$ is decoded to recover the corresponding RGB content $\hat{\mathbf{X}}_{t}^{re}$. The fusion reconstruction loss is defined as:
\begin{equation}
\mathcal{L}_{fus}^{re}
=
\frac{1}{|\mathcal{M}|}
\sum_{t}\sum_{i\in\mathcal{M}_{t}}
\left\|
\mathbf{X}_{t,i}^{r}
-
\hat{\mathbf{X}}_{t,i}^{re}
\right\|_{2}^{2}.
\end{equation}
Similarly, the RGB-Event-Voxel reconstruction loss $\mathcal{L}_{fus}^{rev}$ is computed only on frames with valid Voxel observations. The overall fusion reconstruction loss is defined as:
\begin{equation}
\mathcal{L}_{fus}
=
\mathcal{L}_{fus}^{re}
+
\mathcal{L}_{fus}^{rev}.
\end{equation}
The fusion reconstruction objective therefore encourages both RGB-Event and RGB-Event-Voxel representations to integrate dynamic information from Event and Voxel with the appearance cues provided by RGB observations.

Meanwhile, we employ symmetric contrastive learning to promote shared representation learning across different modalities. Given the $L_2$-normalized RGB and Event representations $\mathbf{z}_i^r$ and $\mathbf{z}_i^e$, their pairwise similarity is computed as
$s_{ij}^{re}=l_s(\mathbf{z}_i^r)^\top\mathbf{z}_j^e$,
where $l_s$ denotes the learnable scaling factor.

The RGB-Event contrastive objective considers both RGB-to-Event and Event-to-RGB directions, treating corresponding observations as positive pairs and other valid observations as negatives. Similarly, RGB-Voxel contrastive learning is performed in both directions. Following the original CM3AE formulation, the overall contrastive objective is expressed as:
\begin{equation}
\mathcal{L}_{con}
=
\mathcal{L}_{r\rightarrow e}
+
\mathcal{L}_{e\rightarrow r}
+
\mathcal{L}_{r\rightarrow v}
+
\mathcal{L}_{v\rightarrow r}.
\end{equation}
To reduce false-negative interference in sequential observations, other frames from the same video are excluded from the negative set. RGB-Event contrastive learning further filters frames with extremely low Event activity, while RGB-Voxel contrastive learning is performed only on frames with valid Voxel observations.

For Event-conditioned RGB image generation, we employ the standard diffusion denoising objective to optimize the Event conditioning representations using a frozen pre-trained diffusion model. Given the Event-conditioned representation $\mathbf{C}_{t}^{e}$, the diffusion loss is formulated as:
\begin{equation}
\mathcal{L}_{diff}
=
\mathbb{E}_{t,\tau,\boldsymbol{\epsilon}}
\left[
\left\|
\boldsymbol{\epsilon}
-
\epsilon_{\theta}
\left(
\mathbf{z}_{t,\tau},
\tau,
\mathbf{C}_{t}^{e}
\right)
\right\|_{2}^{2}
\right],
\end{equation}
where $\boldsymbol{\epsilon}$ denotes the Gaussian noise added to the RGB latent representation and $\epsilon_{\theta}(\cdot)$ denotes the frozen diffusion U-Net. Since the diffusion backbone remains frozen, gradients from $\mathcal{L}_{diff}$ are propagated through the conditioning pathway to guide the Event representations in learning RGB appearance-related information under the visual generative priors of the diffusion model.

For video-level text generation, the projected RGB-Event visual tokens $\mathbf{V}$ and a textual prompt $\mathbf{p}$ are fed into the frozen Llama-3-8B model to predict the corresponding video-level description $\mathbf{y}=\{y_l\}_{l=1}^{L_y}$. The text generation objective is defined as:
\begin{equation}
\mathcal{L}_{text}
=
-\frac{1}{L_y}
\sum_{l=1}^{L_y}
\log
p
\left(
y_l
\mid
\mathbf{V},
\mathbf{p},
y_{<l}
\right).
\end{equation}
Only the target caption tokens contribute to $\mathcal{L}_{text}$, while the visual tokens and textual prompt are excluded from loss computation. The pre-trained language model remains frozen, such that the language semantic prior is transferred to the RGB-Event representations through the learnable visual conditioning pathway.

Based on the above objectives, CM3AE 2.0 is optimized in two stages. In the first stage, we focus on event-aware spatiotemporal masked modeling and cross-modal representation learning. The overall objective is defined as
\begin{equation}
\mathcal{L}_{\mathrm{I}}
=
\lambda_{m}\mathcal{L}_{m}
+
\lambda_{esp}\mathcal{L}_{esp}
+
\lambda_{f}\mathcal{L}_{fus}
+
\lambda_{c}\mathcal{L}_{con},
\end{equation}
where $\lambda_{m}$, $\lambda_{esp}$, $\lambda_{f}$, and $\lambda_{c}$ balance the contributions of different objectives. Spatiotemporal masked reconstruction models temporal relationships across continuous RGB-Event observations, while multimodal fusion reconstruction and cross-modal contrastive learning capture complementary information and shared representations across modalities. Together, these objectives promote unified multimodal spatiotemporal representation learning.

The second stage is initialized from the model learned in Stage 1 and further incorporates visual- and language-level generative semantic supervision. Accordingly, the overall objective is formulated as
\begin{equation}
\mathcal{L}_{\mathrm{II}}
=
\lambda_{m}\mathcal{L}_{m}
+
\lambda_{esp}\mathcal{L}_{esp}
+
\lambda_{f}\mathcal{L}_{fus}
+
\lambda_{d}\mathcal{L}_{diff}
+
\lambda_{t}\mathcal{L}_{text},
\end{equation}
where $\lambda_{d}$ and $\lambda_{t}$ control the contributions of the diffusion and text generation objectives, respectively. Through this two-stage optimization, CM3AE 2.0 first learns unified multimodal spatiotemporal representations and then explores latent cross-modal dependencies by linking Event dynamics with RGB appearance and joint RGB-Event representations with scene-level semantics.

\subsection{Downstream Tasks}  

To comprehensively evaluate the transferability and generalization capability of CM3AE 2.0, we conduct experiments on four representative downstream tasks, including action recognition, object detection, visual object tracking, and event-guided low-light image enhancement. These tasks cover both Event-based unimodal perception and RGB-Event multimodal understanding, spanning temporal recognition, spatial perception, cross-modal fusion, and low-level visual enhancement. For downstream adaptation, the reconstruction and generation branches used during pre-training are discarded, while the pre-trained modality-specific encoders and temporal modeling components relevant to each task are transferred to task-specific networks.

\noindent $\bullet$ \textbf{Task 1. Action Recognition Task.}
Action recognition requires effective modeling of spatial appearance and temporal motion patterns in video sequences. We adopt TimeSformer~\cite{bertasius2021space} as the downstream framework and evaluate CM3AE 2.0 under both Event-only and RGB-Event settings. The pre-trained representations are transferred to the corresponding recognition networks for task-specific fine-tuning. This task evaluates the transferability of the learned multimodal spatiotemporal representations.

\noindent $\bullet$ \textbf{Task 2. Object Detection Task.}
Object detection aims to recognize and localize objects in complex scenes. We adopt ViTDet~\cite{li2022exploring} and SPFD~\cite{wang2026beyond} as the downstream frameworks for Event-only and RGB-Event detection, respectively. The pre-trained representations of CM3AE 2.0 are incorporated into the corresponding detection networks for fine-tuning. This task evaluates their effectiveness in spatial perception and multimodal feature learning.

\noindent $\bullet$ \textbf{Task 3. Visual Object Tracking Task.}
Visual object tracking aims to continuously localize a target throughout a video sequence, requiring robust spatial and temporal representations. We adopt ODTrack~\cite{zheng2024odtrack} and CEUTrack~\cite{tang2025revisiting} for Event-only and RGB-Event tracking, respectively. The pre-trained representations of CM3AE 2.0 are transferred to the corresponding tracking networks for fine-tuning. This task evaluates their transferability to sequential target perception.

\noindent $\bullet$ \textbf{Task 4. Event-guided Low-Light Image Enhancement Task.}
Event-guided low-light image enhancement aims to improve image quality under challenging illumination conditions by leveraging complementary RGB and Event information. We adopt EvLight~\cite{liang2024towards} as the downstream framework and incorporate the pre-trained representations of CM3AE 2.0 for task-specific fine-tuning. This task evaluates the transferability of the learned representations to low-level visual enhancement.

\begin{table*}
\centering
\caption{Comparison with existing pre-trained models across three RGB-Event downstream tasks, including action recognition, object detection, and visual object tracking.}
\label{tab:result_all_1} 

\begin{tabular}{l|c|cc|ccc|ccc}
    \hline \toprule

    \multirow{3}{*}{\raggedright \textbf{Method}} 
    & \multirow{3}{*}{\raggedright \textbf{Dataset}} 
    & \multicolumn{2}{c|}{\textbf{Action Recognition}} 
    & \multicolumn{3}{c|}{\textbf{Object Detection}}  
    & \multicolumn{3}{c}{\textbf{Visual Object Tracking}}   \\
    
    & & \multicolumn{2}{c|}{\textit{(HARDVS Dataset~\cite{wang2024hardvs})}}
      & \multicolumn{3}{c|}{\textit{(DSEC-Det Dataset~\cite{gehrig2024low})}}
      & \multicolumn{3}{c}{\textit{(COESOT Dataset~\cite{tang2025revisiting})}}    \\
    
    \cline{3-10}
    
    & & Top-1 & Top-5 
      & $AP_{[0.5:0.95]}$ & $AP_{0.5}$ & $AP_{0.75}$  
      & SR & PR & NPR  \\
    
    \hline 
    
    Scratch & -   
    & 41.63 & 50.66    
    & 38.6 & 58.7 & 43.5    
    & 38.9 & 44.7 & 46.4   \\
    
    iBOT~\cite{zhouimage} & ImageNet-1K    
    & 45.12 & 51.36   
    & 40.9 & 62.5 & 45.1    
    & 58.3 & 71.3 & 70.3   \\
    
    BEiT v2~\cite{peng2022beit} & ImageNet-1K    
    & 45.01 & 51.76   
    & 41.3 & 63.3 & 46.5    
    & 43.1 & 51.6 & 53.1   \\
    
    EVA~\cite{fang2023eva} & ImageNet-1K    
    & 48.14 & 55.06   
    & 41.1 & 62.5 & 45.7    
    & 58.1 & 70.7 & 70.5    \\
    
    PixMIM~\cite{liupixmim} & ImageNet-1K    
    & 47.17 & 53.44   
    & 42.0 & 62.7 & 47.0  
    & 59.4 & 72.8 & 71.2     \\
    
    MAE~\cite{he2022mae} & ImageNet-1K    
    & 50.23 & 59.65   
    & 41.7 & 62.9 & 46.4    
    & 60.6 & 73.6 & 72.8     \\
    
    \hline 

    MEM~\cite{klenk2024masked} & N-ImageNet    
    & 47.70 & 56.73   
    & 40.6 & 62.2 & 45.4    
    & 54.4 & 62.8 & 64.2   \\
    
    EventBind~\cite{zhou2024eventbind} & WIT     
    & 49.13 & 58.43   
    & 41.7 & 62.9 & 47.1    
    & 56.7 & 68.5 & 67.3   \\
    
    ECDDP~\cite{yang2024event} & E-TartanAir    
    & 50.13 & 59.29   
    & 41.2 & 62.5 & 45.9    
    & 58.3 & 70.5 & 71.6   \\
    
    \hline 

    MAE~\cite{he2022mae} & REV2M    
    & 50.98 & 62.10    
    & 42.2 & 62.7 & 46.7    
    & 61.0 & 73.8 & 73.1   \\
    
    CM3AE~\cite{wu2025cm3ae} & REV2M    
    & 53.40 & 65.82    
    & 43.1 & 64.6 & 48.9    
    & 62.6 & 75.8 & 74.9    \\

    \rowcolor[rgb]{0.95,0.95,0.95} 
    CM3AE 2.0 & REV2M    
    & \textbf{53.95} & \textbf{66.27}    
    & \textbf{44.5} & \textbf{65.4} & \textbf{50.6}   
    & \textbf{63.5} & \textbf{77.0} & \textbf{76.2}     \\
    
    \bottomrule
\end{tabular}
\end{table*}

\begin{table*}
\centering
\caption{Comparison with existing pre-trained models across four event-based downstream tasks, including action recognition, object detection, visual object tracking, and event-guided low-light image enhancement.}
\label{tab:result_all_2} 
\resizebox{\textwidth}{!}{
\begin{tabular}{l|c|cc|ccc|ccc|ccc}
    \hline \toprule
    \multirow{3}{*}{\raggedright \textbf{Method}} 
    & \multirow{3}{*}{\raggedright \textbf{Dataset}} 
    & \multicolumn{2}{c|}{\textbf{Action Recognition}} 
    & \multicolumn{3}{c|}{\textbf{Object Detection}}  
    & \multicolumn{3}{c|}{\textbf{Visual Object Tracking}}   
    & \multicolumn{3}{c}{\textbf{Event-guided Image Enhancement}}\\

    & & \multicolumn{2}{c|}{\textit{(HARDVS Dataset~\cite{wang2024hardvs})}}
      & \multicolumn{3}{c|}{\textit{(EvDET200K Dataset~\cite{wang2025object})}}
      & \multicolumn{3}{c|}{\textit{(EventVOT Dataset~\cite{wang2024event})}}
      & \multicolumn{3}{c}{\textit{(SDE-in Dataset~\cite{liang2024towards})}} \\

    \cline{3-13}

    & & Top-1 & Top-5 
      & $AP_{[0.5:0.95]}$ & $AP_{0.5}$ & $AP_{0.75}$ 
      & SR & PR & NPR 
      & SSIM & PSNR & PSNR* \\
    
    \hline 

    Scratch & -   
    & 20.23 & 37.74    
    & 42.2 & 73.0 & 44.0    
    & 44.8 & 41.8 & 57.9   
    & 70.2 & 20.2 & 23.2 \\

    iBOT~\cite{zhouimage} & ImageNet-1K    
    & 22.99 & 39.99    
    & 46.2 & 76.3 & 49.1    
    & 55.7 & 59.7 & 70.6   
    & 74.2 & 21.6 & 24.1 \\

    BEiT v2~\cite{peng2022beit} & ImageNet-1K    
    & 26.61 & 43.52    
    & 46.6 & 78.4 & 48.4    
    & 51.1 & 52.0 & 65.0   
    & 73.8 & 21.5 & 24.0 \\

    EVA~\cite{fang2023eva} & ImageNet-1K    
    & 36.44 & 50.96   
    & 47.5 & 78.7 & 50.7    
    & 56.0 & 60.3 & 71.1    
    & 74.6 & 21.5 & 24.2 \\

    PixMIM~\cite{liupixmim} & ImageNet-1K    
    & 35.07 & 49.67   
    & 47.3 & 77.5 & 50.7  
    & 55.6 & 59.7 & 70.5     
    & 75.0 & 21.8 & 24.6 \\

    MAE~\cite{he2022mae} & ImageNet-1K    
    & 49.21 & 59.01   
    & 47.4 & 78.1 & 50.9    
    & 56.6 & 60.7 & 71.4     
    & 75.2 & 21.9 & 24.9 \\
    
    \hline 

    MEM~\cite{klenk2024masked} & N-ImageNet    
    & 45.10 & 53.80   
    & 46.5 & 77.5 & 49.1    
    & 51.3 & 53.5 & 65.4   
    & 72.0 & 21.2 & 23.7 \\

    EventBind~\cite{zhou2024eventbind} & WIT     
    & 44.22 & 52.96   
    & 45.0 & 74.9 & 47.7    
    & 49.3 & 49.2 & 63.1   
    & 73.7 & 21.8 & 24.0 \\

    ECDDP~\cite{yang2024event} & E-TartanAir     
    & 49.64 & 56.02   
    & 47.3 & 77.8 & 50.3    
    & 54.3 & 59.6 & 69.0   
    & 75.4 & 22.0 & 24.7 \\

    \hline 

    MAE~\cite{he2022mae} & REV2M    
    & 49.32 & 60.41    
    & 47.5 & 78.0 & 50.3    
    & 56.8 & 60.5 & 71.5   
    & 75.5 & 21.9 & 24.6 \\

    CM3AE~\cite{wu2025cm3ae} & REV2M    
    & 52.45 & 63.76    
    & 49.0 & 80.0 & 52.4    
    & 57.6 & 62.3 & 72.8    
    & 76.0 & 22.2 & 24.7 \\

    \rowcolor[rgb]{0.95,0.95,0.95} 
    CM3AE 2.0 & REV2M    
    & \textbf{53.26} & \textbf{64.58}   
    & \textbf{50.3} & \textbf{81.5} & \textbf{53.1}    
    & \textbf{59.0} & \textbf{63.2} & \textbf{74.1}   
    & \textbf{77.0} & \textbf{22.5} & \textbf{25.3} \\

    \bottomrule
\end{tabular}}
\end{table*}

\begin{figure}
\centering
\includegraphics[width=3.5in]{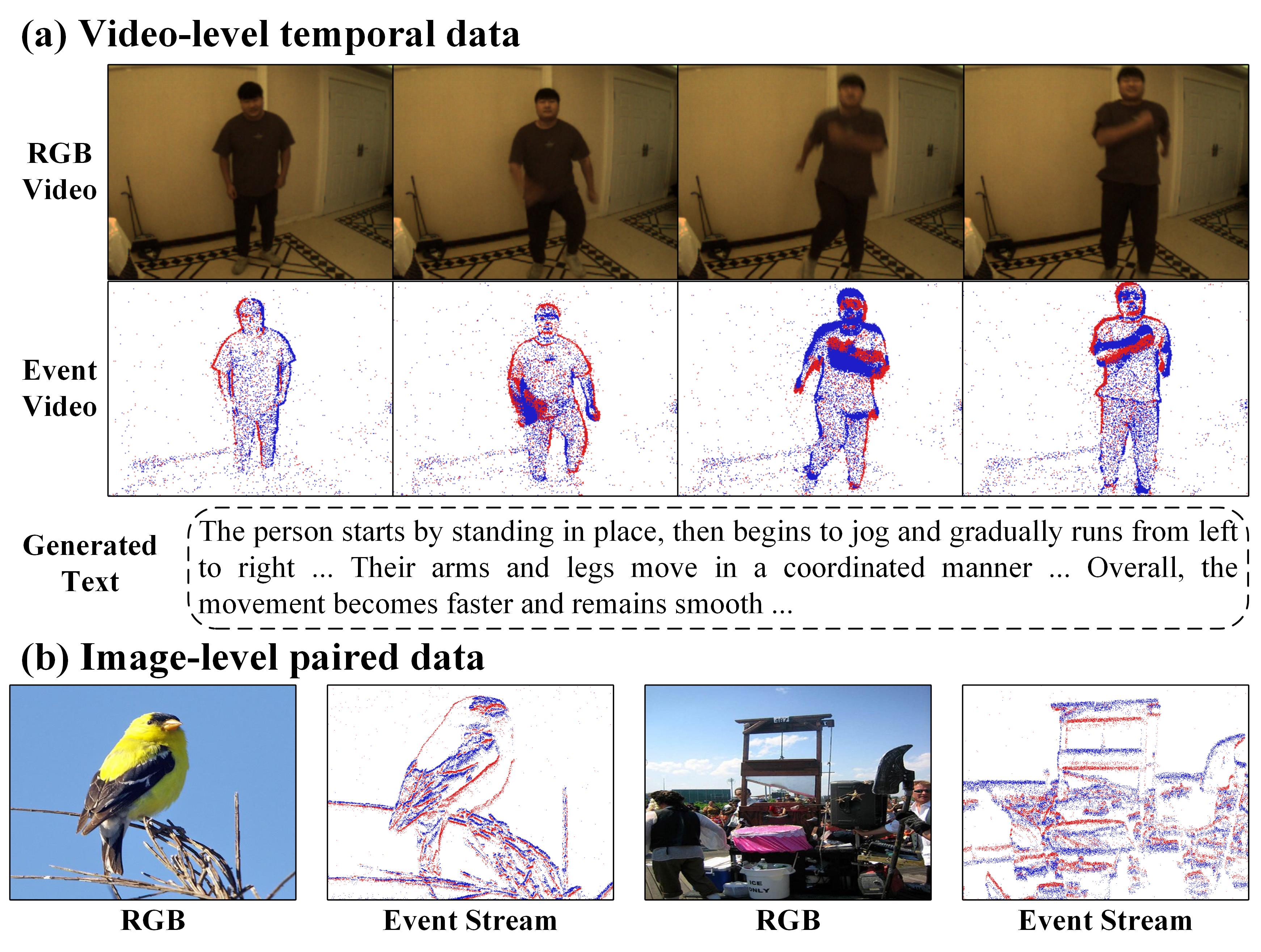}
\caption{Representative samples in our pre-training dataset.} 
\label{fig:dataset} 
\end{figure}

\section{Experiments} \label{sec::experiments}

\subsection{Datasets and Evaluation Metrics}  \label{DSdatasets}
We construct a large-scale RGB-Event dataset REV2M for pre-training and evaluate the proposed method on multiple downstream benchmarks covering action recognition, object detection, visual object tracking, and event-guided low-light image enhancement. In the following, we first introduce the pre-training and downstream datasets, and then describe the evaluation metrics adopted for different tasks.

\noindent $\bullet$ \textbf{Pre-training Dataset.}
To construct a large-scale RGB-Event pre-training dataset, we integrate four publicly available datasets, including HARDVS~\cite{wang2024hardvs}, N-ImageNet~\cite{kim2021nimagenet}, COESOT~\cite{tang2025revisiting}, and VisEvent~\cite{wang2023visevent}. The source video portion contains 66,035 video sequences, while the image-level portion consists of 1,025,012 valid images. These data provide a large-scale and diverse foundation for unified RGB-Event pre-training. Representative examples of the video-level temporal data and image-level paired data are shown in Fig.~\ref{fig:dataset}. To accommodate different data formats, we construct the pre-training samples in a unified sequential form. For video data with continuous temporal information, the original videos are divided into fixed-length RGB-Event clips, resulting in 292,364 candidate clips as pre-training samples after filtering. For image-level data without genuine temporal continuity, the images are organized into fixed-length pseudo-sequences, with each sample containing four images, and temporal interaction is disabled during training. In total, the 1,025,012 images are organized into 256,253 image-level pseudo-sequence samples. After the above data processing and filtering procedures, the resulting pre-training dataset contains 548,617 training samples in total, including 292,364 video clip samples and 256,253 image-level pseudo-sequence samples. This large-scale pre-training corpus supports unified multimodal spatiotemporal representation learning by modeling temporal dynamics and exploiting complementary information and shared features across modalities. To prevent data leakage in downstream evaluation, we ensure that the pre-training samples do not overlap with the validation and test sequences of the downstream benchmarks.

\begin{table*}[t] 
\centering 
\caption{Ablation results of the proposed pre-training framework on four representative downstream tasks.} 
\label{tab:main_ablation} 
\resizebox{\textwidth}{!}{ 
\begin{tabular}{l|c|cc|ccc|ccc|ccc} 
\hline \toprule 

\multirow{3}{*}{\textbf{Method}}  
& \multirow{3}{*}{\textbf{Stage}} 
& \multicolumn{2}{c|}{\textbf{Action Recognition}} 
& \multicolumn{3}{c|}{\textbf{Object Detection}} 
& \multicolumn{3}{c|}{\textbf{Visual Object Tracking}} 
& \multicolumn{3}{c}{\textbf{Event-guided Image Enhancement}} \\ 

& 
& \multicolumn{2}{c|}{\textit{(HARDVS Dataset)}}
& \multicolumn{3}{c|}{\textit{(DSEC-Det Dataset)}}
& \multicolumn{3}{c|}{\textit{(COESOT Dataset)}}
& \multicolumn{3}{c}{\textit{(SDE-in Dataset)}} \\

\cline{3-13} 
 
&  
& Top-1 & Top-5 
& $AP_{[0.5:0.95]}$ & $AP_{0.5}$ & $AP_{0.75}$ 
& SR & PR & NPR 
& SSIM & PSNR & PSNR$^{*}$ \\ 

\hline 
Baseline 
& 1 
&49.29  &59.99 
&41.1  &62.7  &46.3 
&58.7  &71.4  &70.7 
&72.9  &21.5  &24.1  \\ 
\hline
\begin{tabular}[c]{@{}l@{}}
+ Multimodal Fusion Reconstruction \\
\phantom{+ }and Cross-modal Contrastive Learning
\end{tabular}
& 1 
&50.51  &61.39 
&42.4 &63.6  &47.8 
&60.4  &72.9  &72.6 
&73.9  &21.9  &24.2  \\ 
\hline
+ Event-guided Masking
& 1 
&51.47  &62.41 
&42.8  &64.2  &48.1 
&61.0  &73.7  &73.4  
&74.5  &22.2  &24.3   \\ 
\hline
+ Spatiotemporal Masked Reconstruction
& 1 
&52.76  &64.86 
&43.8  &65.1  &49.0  
&62.3  &75.5  &74.7 
&75.1  &22.3  &24.6  \\ 
\hline
+ Event-specific Reconstruction Loss
& 1 
&53.28  &65.63  
&44.1  &65.2  &49.5  
&62.8  &75.9  &75.1 
&75.6  &22.3  &24.7  \\ 
\hline 
+ Event-conditioned Image Generation 
& 2 
&53.64  &66.01  
&44.3  &65.4  &50.2  
&63.3  &76.6  &75.9 
&76.8  &\textbf{22.6}  &25.1  \\ 
\hline
\rowcolor[rgb]{0.95,0.95,0.95} 
+ Video-level Text Generation \textbf{(Full Model)} 
& 2 
& \textbf{53.95} & \textbf{66.27} 
& \textbf{44.5} & \textbf{65.4} & \textbf{50.6} 
& \textbf{63.5} & \textbf{77.0} & \textbf{76.2} 
& \textbf{77.0} & 22.5 & \textbf{25.3} \\ 
\bottomrule 
\end{tabular} 
} 
\end{table*}

\begin{figure*}
\centering
\includegraphics[width=\textwidth]{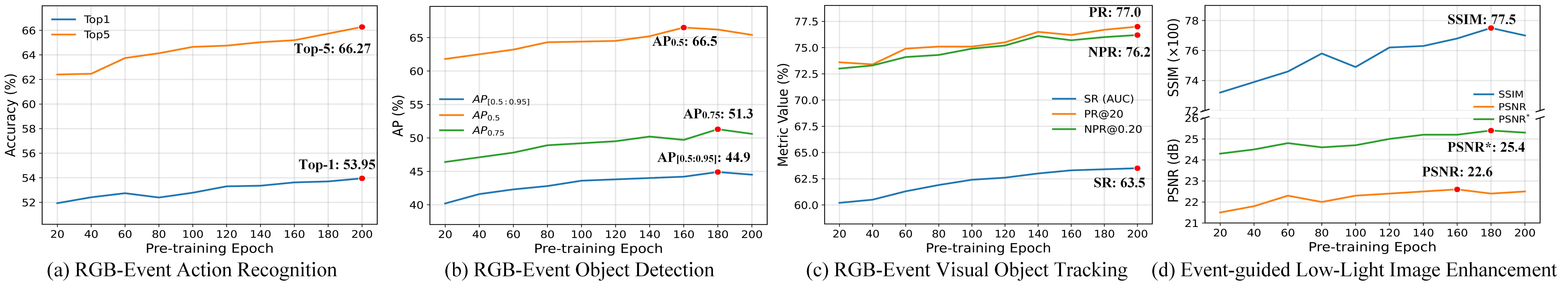}
\caption{The impact of pre-training epochs on experimental results across different downstream tasks.}
\label{fig:epoch} 
\end{figure*}

\noindent $\bullet$ \textbf{Downstream Datasets.}
We evaluate the transferability of the learned representations on multiple downstream benchmarks, covering action recognition, object detection, visual object tracking, and event-guided low-light image enhancement. The datasets used in our experiments are introduced as follows.

\textbf{\textit{HARDVS}}~\cite{wang2024hardvs} is a large-scale benchmark for human action recognition using event data. It contains 107,646 event sequences covering 300 action categories, with diverse action types and scene variations. We evaluate CM3AE 2.0 on HARDVS under both Event-only and RGB-Event settings. The Event-only setting uses event sequences as input, while the RGB-Event setting jointly uses the corresponding RGB observations and Event data. Following the protocol in the original paper, the dataset is divided into training, validation, and testing sets with ratios of 60\%, 10\%, and 30\%, respectively.

\textbf{\textit{DSEC-Det}}~\cite{gehrig2024low} is an RGB-Event object detection dataset for autonomous driving scenarios, containing synchronized image and event data collected in complex traffic environments such as urban roads. The dataset consists of 60 sequences and eight traffic-related object categories, providing 70,379 annotated frames and 390,118 bounding boxes. According to the official split, the training, validation, and testing sets contain 41, 6, and 13 sequences, respectively.

\textbf{\textit{EvDET200K}}~\cite{wang2025object} is a large-scale event-based object detection dataset captured using a high-resolution Prophesee EVK4-HD event camera. It contains diverse real-world scenes and targets under different motion conditions. The dataset includes 10,054 event samples covering 10 object categories and provides more than 200K bounding-box annotations. 

\textbf{\textit{COESOT}}~\cite{tang2025revisiting} is a large-scale multimodal dataset for RGB-Event single-object tracking, where RGB images and event data are synchronously captured using an event camera. It covers diverse indoor and outdoor environments as well as challenging conditions such as low illumination, fast motion, occlusion, and scale variation. The dataset contains 1,354 RGB-Event video sequences covering 90 target categories, with 827 sequences for training and 527 sequences for testing.

\textbf{\textit{EventVOT}}~\cite{wang2024event} is a high-resolution benchmark for event-based visual object tracking. Compared with existing low-resolution event tracking datasets, EventVOT provides event data at a resolution of $1280\times720$ and includes diverse targets such as pedestrians, vehicles, and UAVs under complex motion and scene variations. The dataset contains 1,141 event videos covering 19 target categories and 14 challenging attributes. The training, validation, and testing sets contain 841, 18, and 282 videos, respectively.

\textbf{\textit{SDE-in}}~\cite{liang2024towards} is the indoor subset of the SDE dataset and is mainly used for event-guided low-light image enhancement. SDE is captured using a DAVIS346 camera in real-world low-light environments and provides spatially and temporally aligned low-light images, normal-light images, and event data. The complete dataset contains 91 sequences, including 43 indoor and 48 outdoor sequences. The original SDE dataset uses 76 sequences for training and 15 sequences for testing. In our experiments, we follow the original setting and use its indoor subset, SDE-in, for evaluation.

\noindent $\bullet$ \textbf{Evaluation Metrics.}
For action recognition, we adopt Top-1 and Top-5 classification accuracies as the evaluation metrics. Top-1 accuracy measures the proportion of samples for which the class with the highest predicted probability matches the ground-truth label, while Top-5 accuracy measures the proportion of samples for which the ground-truth label is included among the five classes with the highest predicted probabilities.
For object detection, we use the COCO-style Average Precision (AP), including $AP_{0.5}$, $AP_{0.75}$, and $AP_{[0.5:0.95]}$. Specifically, $AP_{0.5}$ and $AP_{0.75}$ denote the average precision at IoU thresholds of 0.5 and 0.75, respectively, while $AP_{[0.5:0.95]}$ represents the average AP over multiple IoU thresholds ranging from 0.5 to 0.95 with a step size of 0.05.
For visual object tracking, we adopt Success Rate (SR), Precision Rate (PR), and Normalized Precision Rate (NPR). SR mainly evaluates the overlap between the predicted and ground-truth bounding boxes, PR measures the localization accuracy based on the center distance between the predicted and ground-truth targets, and NPR further normalizes the center localization error to reduce the influence of variations in target scale and image resolution.
For event-guided low-light image enhancement, we employ Peak Signal-to-Noise Ratio (PSNR), brightness-adjusted PSNR ($\mathrm{PSNR}^{*}$), and Structural Similarity Index Measure (SSIM). PSNR evaluates the pixel-level reconstruction quality between the enhanced image and the reference image, while $\mathrm{PSNR}^{*}$ further compensates for global brightness differences. SSIM measures the similarity between two images in terms of luminance, contrast, and structural information. Higher values indicate better performance for all the above metrics.

\begin{figure*}
\centering
\includegraphics[width=\textwidth]{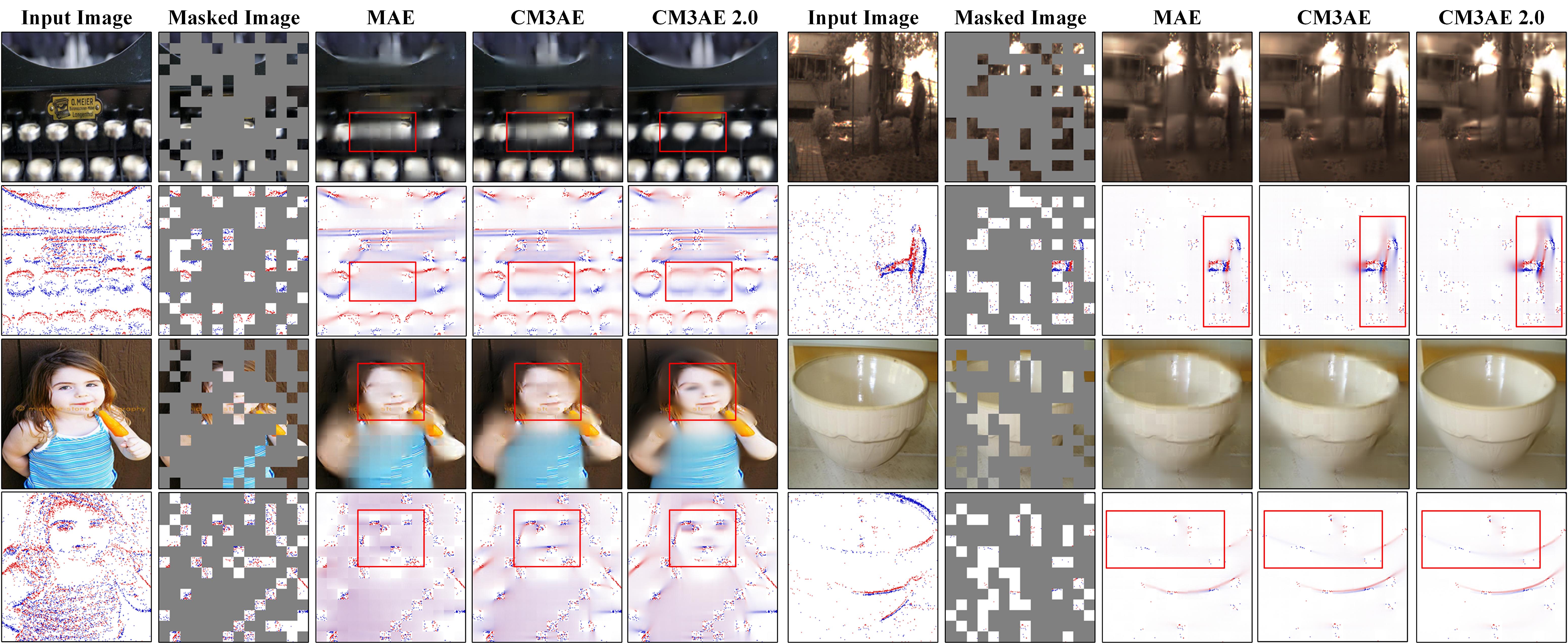}
\caption{Visualization of masked image reconstruction results using MAE, CM3AE, and CM3AE 2.0.}
\label{fig:reconst} 
\end{figure*}

\subsection{Implementation Details} \label{impleDetails} 
During the pre-training phase, we adopt ViT-B/16 as the visual backbone and train CM3AE 2.0 using a two-stage pre-training strategy. All experiments are implemented in Python based on PyTorch~\cite{paszke2019pytorch}, with training conducted in BF16 precision. In the first stage, the model is trained for 100 epochs using the AdamW~\cite{adamW} optimizer. The base and minimum learning rates are set to $2.5\times10^{-4}$ and $1\times10^{-6}$, respectively, while the weight decay is gradually increased from 0.04 to 0.4. A 10-epoch learning-rate warm-up is employed. The batch size is set to 45 per GPU, resulting in an effective batch size of 180 across four GPUs. The masking ratio is set to 0.75, the temporal interaction interval is set to 4, and the guided pairwise masking strategy is adopted with a noise ratio of 0.1. In the second stage, the model is initialized with the weights obtained from the first stage and further trained for 100 epochs. AdamW is again adopted as the optimizer, with the base and minimum learning rates set to $5\times10^{-5}$ and $1\times10^{-6}$, respectively. The weight decay is gradually increased from 0.04 to 0.4, and a 2-epoch learning-rate warm-up is employed. The batch size is set to 24 per GPU, corresponding to an effective batch size of 96 across four GPUs. All pre-training experiments are conducted on four NVIDIA A800 GPUs, and the complete two-stage pre-training process takes approximately 240 hours.

During the fine-tuning stage, task-specific training configurations are adopted for different downstream tasks. For human action recognition on HARDVS, the model is trained for 15 epochs with a learning rate of 0.008 and a batch size of 7. For Event-based object detection on EvDET200K, the model is trained for 15 epochs with a learning rate of 0.0001 and a batch size of 8. For RGB-Event multimodal object detection on DSEC-Det, the model is trained for 120K iterations using the AdamW optimizer, with an initial learning rate of $1\times10^{-4}$ and a total batch size of 2. For Event-based visual object tracking on EventVOT, the model is fine-tuned for 200 epochs with a learning rate of 0.0001 and a batch size of 16. For RGB-Event visual object tracking on COESOT, the model is trained for 50 epochs with a learning rate of 0.0001 and a batch size of 16. For event-guided low-light image enhancement on SDE-in, the model is trained for 80 epochs using the Adam optimizer, with a learning rate of $1.5\times10^{-4}$ and a batch size of 6. The training configurations for the downstream tasks largely follow those of their corresponding baseline methods. All downstream experiments are conducted on NVIDIA RTX 4090 GPUs.

\begin{figure*}
\centering
\includegraphics[width=\textwidth]{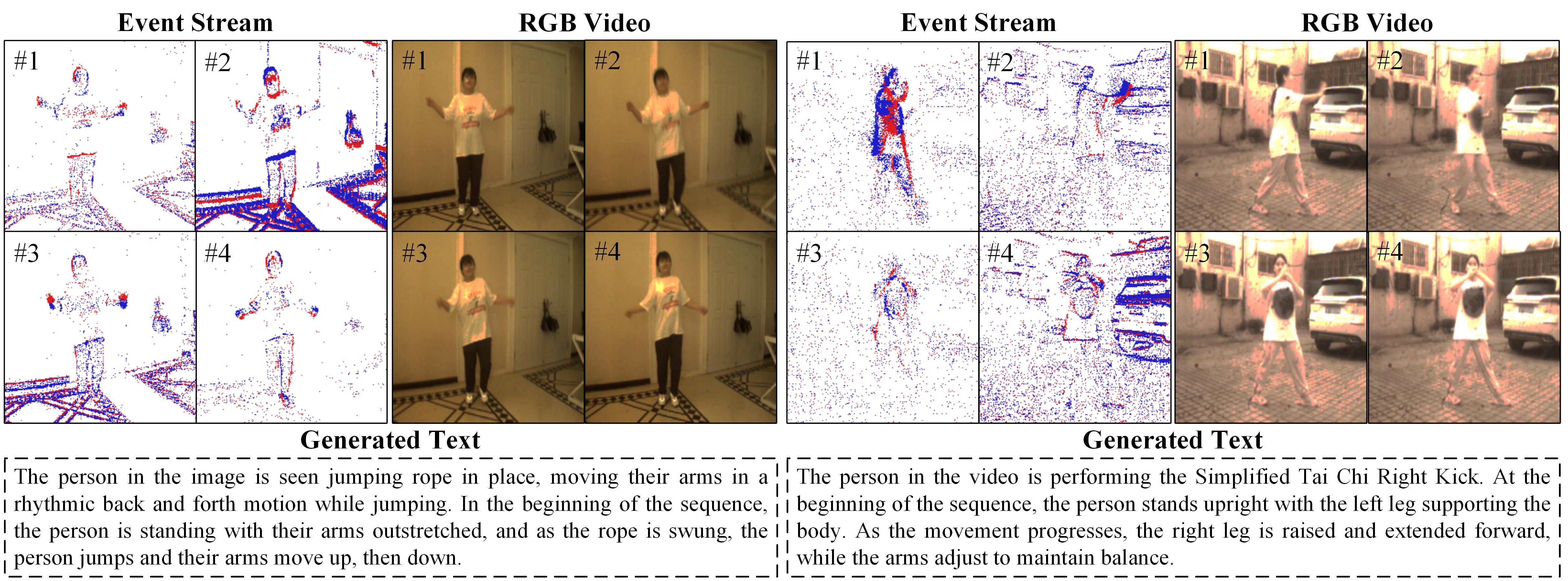}
\caption{Qualitative visualization of video-level text generation.}
\label{fig:txt_generation}
\end{figure*}

\begin{figure}
\centering
\includegraphics[width=3.5in]{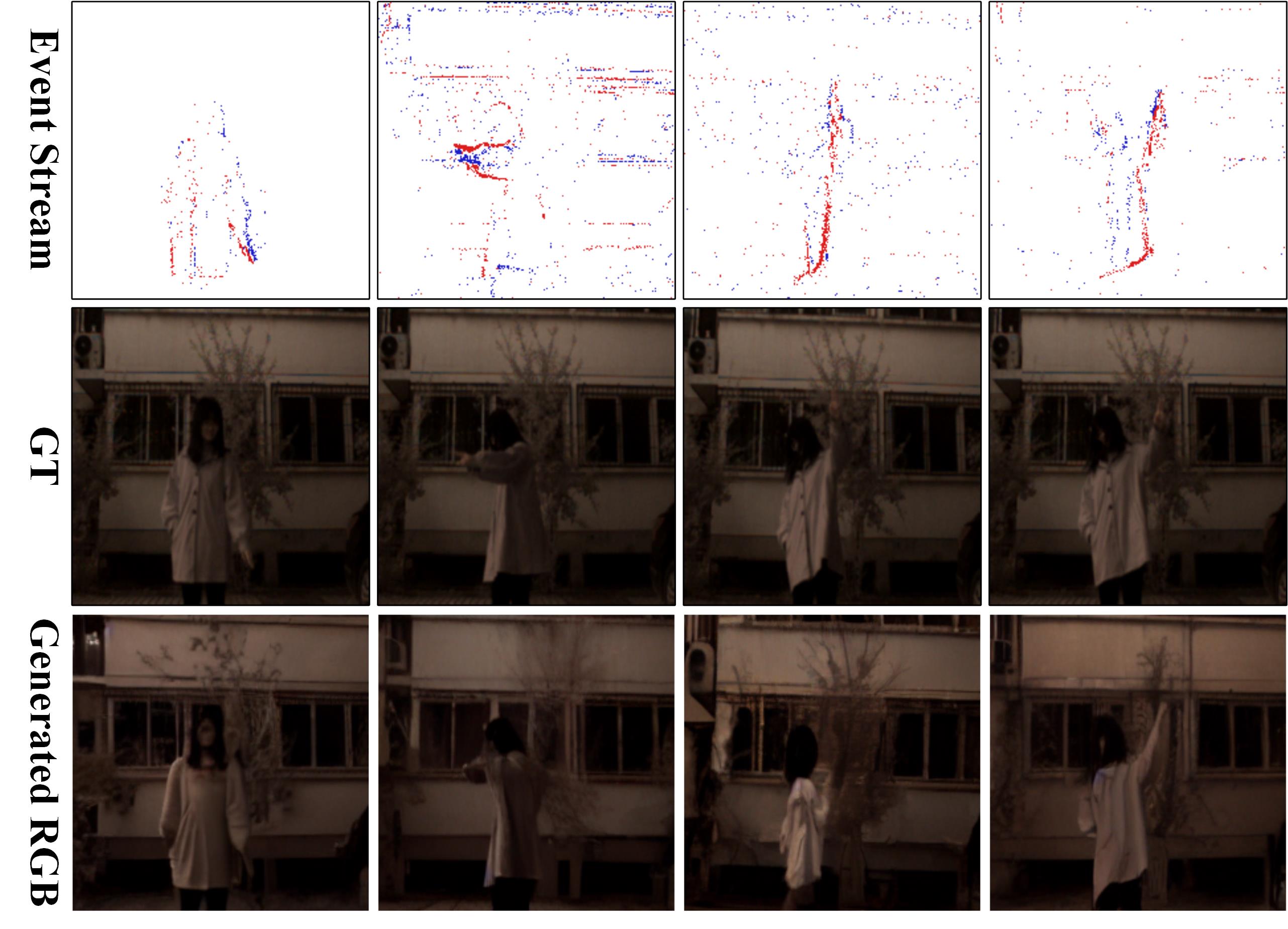}
\caption{Qualitative visualization of RGB images generated from Event-conditioned visual representations by CM3AE 2.0.}
\label{fig:image_generation}
\end{figure}

\subsection{Comparison with State-of-the-Art Algorithms}   \label{compSOTA}

We systematically evaluate the transferability of CM3AE 2.0 across seven downstream evaluation settings covering four representative tasks under Event-only and RGB-Event configurations. As shown in Table~\ref{tab:result_all_1} and Table~\ref{tab:result_all_2}, the compared methods mainly fall into two categories. The first category consists of general-purpose visual pre-training models trained on large-scale RGB images, including iBOT~\cite{zhouimage}, BEiT v2~\cite{peng2022beit}, EVA~\cite{fang2023eva}, PixMIM~\cite{liupixmim}, and MAE~\cite{he2022mae}. The second category includes pre-training methods specifically designed for Event data, including MEM~\cite{klenk2024masked}, EventBind~\cite{zhou2024eventbind}, and ECDDP~\cite{yang2024event}. In addition, we adopt MAE pre-trained on REV2M~\cite{wu2025cm3ae} as a baseline under the same pre-training data setting, and further compare our method with CM3AE~\cite{wu2025cm3ae} proposed in our previous conference version.

\begin{figure*}
\centering
\includegraphics[width=\textwidth]{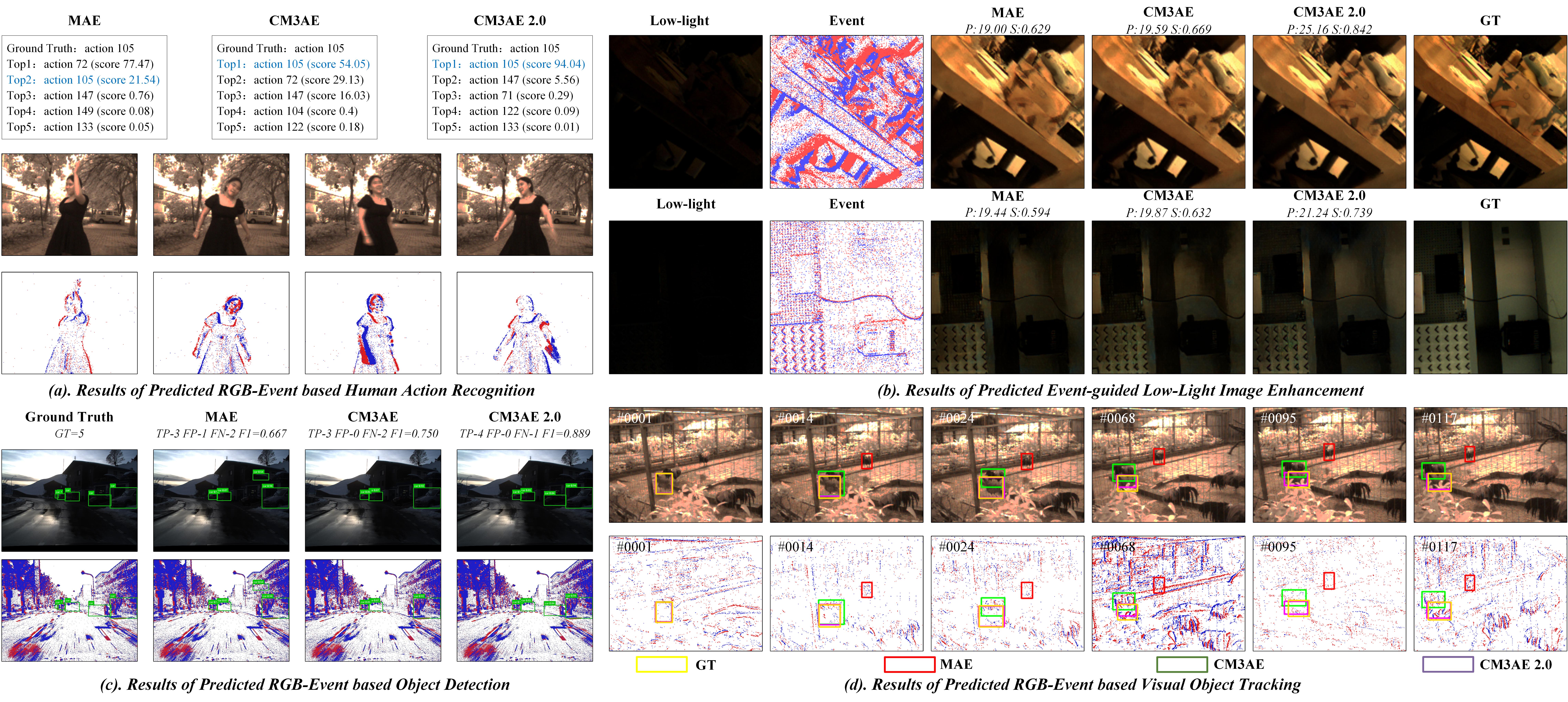}
\caption{Visualization of the experimental results of each downstream task.}
\label{fig:downstream_vis} 
\end{figure*}

\noindent $\bullet$ \textbf{Results of Action Recognition Task.} 
On HARDVS, CM3AE 2.0 achieves $53.95\%$ Top-1 and $66.27\%$ Top-5 accuracy under the RGB-Event setting. Compared with MAE, a widely used general-purpose visual pre-training baseline, CM3AE 2.0 improves Top-1 accuracy by $3.72\%$ and also outperforms ECDDP. Under the same REV2M pre-training data setting, MAE achieves $50.98\%$ Top-1 accuracy, while CM3AE and CM3AE 2.0 further improve it to $53.40\%$ and $53.95\%$, respectively. A consistent trend is also observed under the Event-only setting, where CM3AE 2.0 achieves $53.26\%$ Top-1 accuracy, outperforming MAE, ECDDP, and CM3AE. These results demonstrate the transferability of the learned representations to action recognition under both Event-only and RGB-Event settings.

\noindent $\bullet$ \textbf{Results of Object Detection Task.} 
On the RGB-Event object detection task on DSEC-Det, CM3AE 2.0 achieves $44.5\%$, $65.4\%$, and $50.6\%$ in $AP_{[0.5:0.95]}$, $AP_{0.5}$, and $AP_{0.75}$, respectively. In particular, its $AP_{[0.5:0.95]}$ exceeds that of the strong general-purpose pre-training model PixMIM by $2.5\%$ and that of MAE pre-trained on the same REV2M dataset by $2.3\%$. On the Event-only EvDET200K dataset, CM3AE 2.0 further achieves $50.3\%$ $AP_{[0.5:0.95]}$, outperforming methods such as MAE and ECDDP, as well as CM3AE with $49.0\%$. These results demonstrate the transferability of the learned representations to both Event-only and RGB-Event object detection, supporting their effectiveness in spatial perception and multimodal feature learning.

\noindent $\bullet$ \textbf{Results of Visual Object Tracking Task.}
On COESOT, CM3AE 2.0 achieves $63.5\%$ SR, $77.0\%$ PR, and $76.2\%$ NPR under the RGB-Event setting. Compared with MAE pre-trained on ImageNet-1K, the three metrics are improved by $2.9\%$, $3.4\%$, and $3.4\%$, respectively, while consistent improvements are also obtained over MAE pre-trained on REV2M. On EventVOT, CM3AE 2.0 further achieves $59.0\%$ SR, $63.2\%$ PR, and $74.1\%$ NPR, outperforming MAE, ECDDP, and CM3AE. These results further demonstrate the effectiveness of unified multimodal spatiotemporal representation learning in capturing temporal relationships and supporting robust sequential visual perception.

\begin{table}[t]
\centering
\caption{Effect of different masking ratios on action recognition.}
\label{tab:mask_ratio}
\begin{tabular}{c|cc|cc}
\hline \toprule
\multirow{3}{*}{\textbf{Mask Ratio}}
& \multicolumn{4}{c}{\textbf{Action Recognition}} \\
\cline{2-5}

& \multicolumn{2}{c|}{\textbf{RGB+Event}}
& \multicolumn{2}{c}{\textbf{Event}} \\
\cline{2-5}

& Top-1 & Top-5 & Top-1 & Top-5 \\
\hline

0.50 &52.62  &64.67  &52.28  &63.83  \\
0.75 &53.95  &66.27  &53.26  &64.58  \\ 
0.90 &51.78  &62.83  &51.03  &61.57  \\

\bottomrule
\end{tabular}
\end{table}

\noindent $\bullet$ \textbf{Results of Event-guided Low-Light Image Enhancement Task.}
On SDE-in, CM3AE 2.0 achieves an SSIM of $77.0\%$, a PSNR of 22.5 dB, and a PSNR$^{*}$ of 25.3 dB. Compared with the general-purpose visual pre-training model MAE, all three metrics are improved, while consistent advantages are also observed over the Event-oriented pre-training method ECDDP and MAE pre-trained on REV2M. Moreover, compared with CM3AE, CM3AE 2.0 improves SSIM from $76.0\%$ to $77.0\%$, PSNR from 22.2 dB to 22.5 dB, and PSNR$^{*}$ from 24.7 dB to 25.3 dB. These results demonstrate that the learned representations effectively exploit complementary RGB-Event information for low-light image enhancement, indicating their transferability to low-level visual perception.

\begin{table}[t]
\centering
\caption{Effect of different numbers of input frames on action recognition.}
\label{tab:num_frames}
\begin{tabular}{c|cc|cc}
\hline \toprule
\multirow{3}{*}{\textbf{Frames}}
& \multicolumn{4}{c}{\textbf{Action Recognition}} \\
\cline{2-5}

& \multicolumn{2}{c|}{\textbf{RGB-Event}}
& \multicolumn{2}{c}{\textbf{Event}} \\
\cline{2-5}

& Top-1 & Top-5 & Top-1 & Top-5 \\
\hline

2 &52.13  &64.85  &52.27  &63.74  \\
4 &53.95  &66.27  &53.26  &64.58  \\
8 &54.03  &66.14  &53.19  &64.52  \\

\bottomrule
\end{tabular}
\end{table}

\subsection{Component Analysis} \label{componentAnalysis}

To investigate the contributions of the ESMR and HCG strategies, we progressively evaluate their components, as shown in Table~\ref{tab:main_ablation}. The baseline achieves $49.29\%$ Top-1 accuracy on action recognition, $41.1\%$ $AP_{[0.5:0.95]}$ on object detection, and $58.7\%$ SR on visual object tracking. Introducing multimodal fusion reconstruction and cross-modal contrastive learning improves performance across downstream tasks, supporting the benefits of exploiting complementary information and learning shared features across modalities. Event-guided masking further improves the learned representations by emphasizing dynamically salient regions. Incorporating spatiotemporal masked reconstruction increases the Top-1 accuracy from $51.47\%$ to $52.76\%$ and the tracking SR from $61.0\%$ to $62.3\%$, highlighting the importance of modeling temporal dynamics in continuous observations. The event-specific reconstruction loss further improves performance by supervising event occurrence and polarity prediction.

In Stage 2, event-conditioned RGB image generation further improves performance, particularly on low-light image enhancement, where SSIM increases from $75.6\%$ to $76.8\%$. Incorporating video-level text generation yields the full CM3AE 2.0 model, further increasing SSIM to $77.0\%$ and improving performance on other downstream tasks. Compared with the baseline, the full model improves Top-1 action recognition accuracy from $49.29\%$ to $53.95\%$, detection $AP_{[0.5:0.95]}$ from $41.1\%$ to $44.5\%$, tracking SR from $58.7\%$ to $63.5\%$, and low-light enhancement SSIM from $72.9\%$ to $77.0\%$. These results demonstrate that the ESMR strategy enables unified multimodal spatiotemporal representation learning, while the HCG strategy further explores latent cross-modal dependencies through visual and language generation.

\begin{figure*}
\centering
\includegraphics[width=\textwidth]{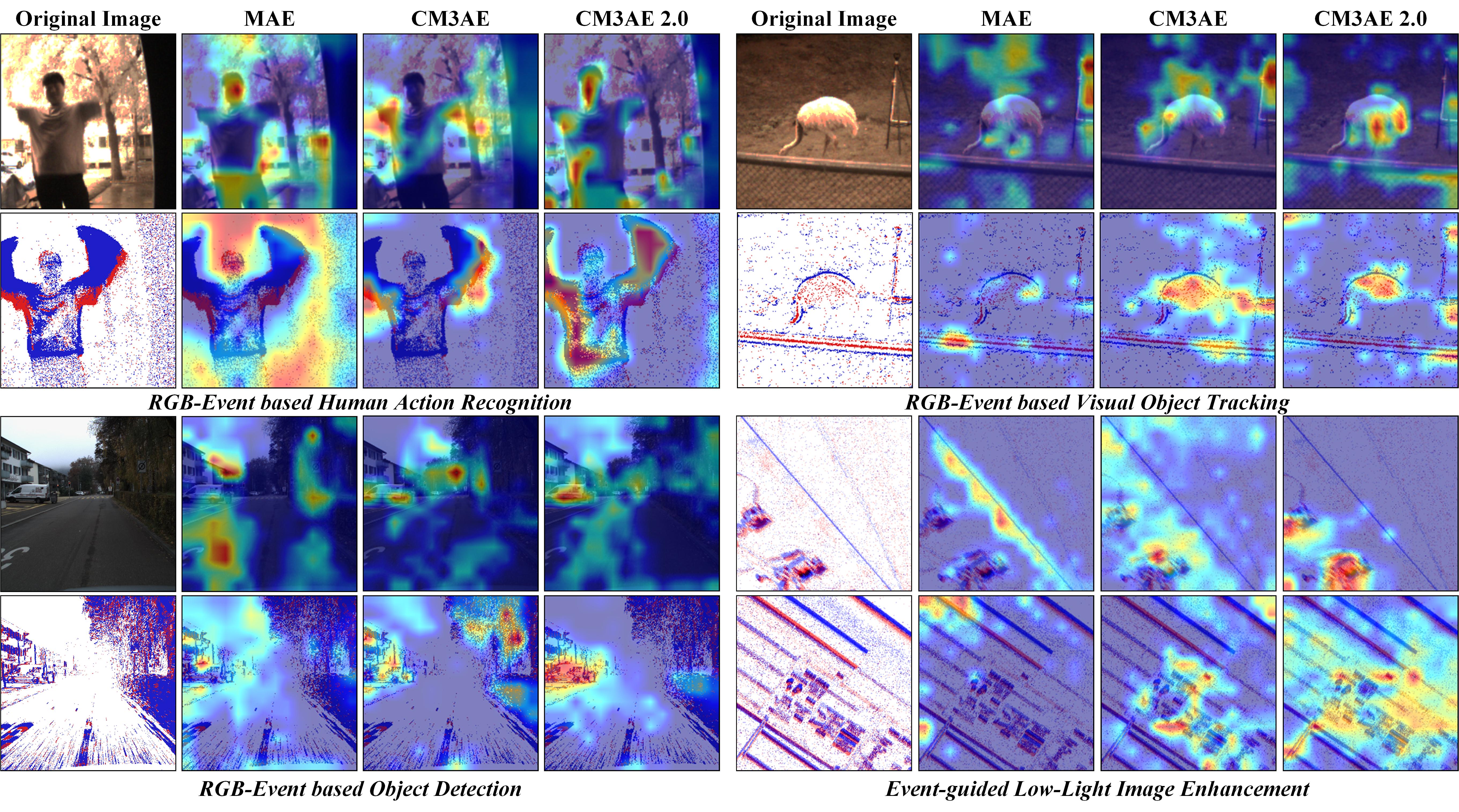}
\caption{Visualization of attention maps on different downstream tasks.}
\label{fig:attentionmaps} 
\end{figure*}

\begin{table*}[t]
\centering
\caption{Few-shot transfer results of different pre-trained models on multiple downstream tasks using 10\% of the labeled training data for fine-tuning.}
\label{tab:few_shot_results}

\footnotesize
\renewcommand{\arraystretch}{1.0}
\setlength{\tabcolsep}{2.5pt}

\begin{tabular}{
>{\raggedright\arraybackslash}p{1.55cm}|
>{\centering\arraybackslash}p{2.35cm}|
*{2}{>{\centering\arraybackslash}p{1.00cm}}|
>{\centering\arraybackslash}p{1.30cm}
>{\centering\arraybackslash}p{0.85cm}
>{\centering\arraybackslash}p{0.85cm}|
*{3}{>{\centering\arraybackslash}p{1.00cm}}|
*{3}{>{\centering\arraybackslash}p{1.00cm}}
}
\toprule

\multirow{3}{*}{\textbf{Method}}
& \multirow{3}{*}{\textbf{Pre-training Data}}
& \multicolumn{2}{c|}{\textbf{Action Recognition}}
& \multicolumn{3}{c|}{\textbf{Object Detection}}
& \multicolumn{3}{c|}{\textbf{Visual Object Tracking}}
& \multicolumn{3}{c}{\textbf{Low-Light Enhancement}}
\\

&
& \multicolumn{2}{c|}{\textit{(HARDVS Dataset)}}
& \multicolumn{3}{c|}{\textit{(DSEC-Det Dataset)}}
& \multicolumn{3}{c|}{\textit{(COESOT Dataset)}}
& \multicolumn{3}{c}{\textit{(SDE-in Dataset)}}
\\

\cline{3-13}

&
& Top-1 & Top-5
& $AP_{[0.5:0.95]}$ & $AP_{0.5}$ & $AP_{0.75}$
& SR & PR & NPR
& SSIM & PSNR & PSNR$^{*}$
\\

\midrule

Scratch
& -
& 5.63 & 14.52
& 17.1 & 29.2 & 17.2
& 26.3 & 26.9 & 30.1
& 59.4 & 19.9 & 20.1
\\

MAE~\cite{he2022mae}
& ImageNet-1K
& 6.65 & 18.04
& 21.1 & 34.2 & 21.4
& 40.1 & 48.7 & 50.7
& 61.3 & 20.2 & 21.3
\\

MAE~\cite{he2022mae}
& REV2M
& 33.31 & 49.01
& 23.1 & 35.8 & 23.7
& 41.5 & 50.8 & 52.3
& 63.2 & 20.7 & 21.7
\\

CM3AE~\cite{wu2025cm3ae}
& REV2M
& 43.68 & 57.23
& 26.6 & 39.8 & 26.1
& 43.9 & 55.5 & 56.0
& 63.8 & 21.1 & 22.2
\\

\rowcolor[rgb]{0.95,0.95,0.95}
CM3AE 2.0
& REV2M
& \textbf{44.96} & \textbf{59.78}
& \textbf{28.4} & \textbf{42.1} & \textbf{27.9}
& \textbf{45.7} & \textbf{59.1} & \textbf{57.9}
& \textbf{69.4} & \textbf{22.1} & \textbf{22.8}
\\

\bottomrule
\end{tabular}

\vspace{2.2mm}

\begin{tabular}{
>{\raggedright\arraybackslash}p{1.55cm}|
>{\centering\arraybackslash}p{2.35cm}|
*{2}{>{\centering\arraybackslash}p{1.52cm}}|
*{3}{>{\centering\arraybackslash}p{1.52cm}}|
*{3}{>{\centering\arraybackslash}p{1.52cm}}
}
\toprule

\multirow{3}{*}{\textbf{Method}}
& \multirow{3}{*}{\textbf{Pre-training Data}}
& \multicolumn{2}{c|}{\textbf{Action Recognition}}
& \multicolumn{3}{c|}{\textbf{Object Detection}}
& \multicolumn{3}{c}{\textbf{Visual Object Tracking}}
\\

&
& \multicolumn{2}{c|}{\textit{(HARDVS Dataset)}}
& \multicolumn{3}{c|}{\textit{(EvDET200K Dataset)}}
& \multicolumn{3}{c}{\textit{(EventVOT Dataset)}}
\\

\cline{3-10}

&
& Top-1 & Top-5
& $AP_{[0.5:0.95]}$ & $AP_{0.5}$ & $AP_{0.75}$
& SR & PR & NPR
\\

\midrule

Scratch
& -
& 2.40 & 8.06
& 26.7 & 47.3 & 26.9
& 25.4 & 20.0 & 30.7
\\

MAE~\cite{he2022mae}
& ImageNet-1K
& 9.31 & 24.63
& 33.3 & 60.2 & 33.3
& 37.0 & 35.5 & 45.9
\\

MAE~\cite{he2022mae}
& REV2M
& 16.94 & 37.27
& 34.2 & 62.1 & 34.9
& 39.2 & 39.8 & 48.1
\\

CM3AE~\cite{wu2025cm3ae}
& REV2M
& 32.34 & 52.18
& 37.7 & 65.9 & 38.7
& 41.1 & 42.7 & 50.5
\\

\rowcolor[rgb]{0.95,0.95,0.95}
CM3AE 2.0
& REV2M
& \textbf{34.52} & \textbf{55.41}
& \textbf{38.3} & \textbf{67.1} & \textbf{39.6}
& \textbf{43.6} & \textbf{44.2} & \textbf{51.9}
\\

\bottomrule
\end{tabular}

\end{table*}

\subsection{Ablation Study} \label{ablationStudy}

\noindent $\bullet$ \textbf{Effects of Pre-training Epochs.} 
We first investigate the influence of the number of pre-training epochs on different downstream tasks. As shown in Fig.~\ref{fig:epoch}, the performance of action recognition, object detection, visual object tracking, and low-light image enhancement generally improves as pre-training progresses. Although slight fluctuations are observed for some metrics, the overall performance continues to improve and gradually becomes stable in the later stage of pre-training. This trend indicates that sufficient pre-training is important for learning unified multimodal spatiotemporal representations and exploring cross-modal dependencies.

\noindent $\bullet$ \textbf{Effects of Masking Ratio.}
Table~\ref{tab:mask_ratio} reports the results with different masking ratios. When the masking ratio increases from 0.50 to 0.75, the RGB-Event Top-1 accuracy improves from $52.62\%$ to $53.95\%$, while the Event-only Top-1 accuracy increases from $52.28\%$ to $53.26\%$. However, further increasing the masking ratio to 0.90 causes noticeable performance degradation under both settings. These results indicate that a moderate masking ratio of 0.75 provides a better balance between reconstruction difficulty and the amount of visible contextual information, and is therefore adopted as the default setting.

\noindent $\bullet$ \textbf{Effects of the Number of Input Frames.}
We further evaluate the effect of different numbers of input frames in Table~\ref{tab:num_frames}. Increasing the number of frames from 2 to 4 consistently improves both RGB-Event and Event-only action recognition. When the number of frames is further increased to 8, the RGB-Event Top-1 accuracy only slightly increases from $53.95\%$ to $54.03\%$, while the other metrics show no further improvement. Considering both performance and computational cost, four input frames provide a better overall trade-off and are adopted in our experiments.

\noindent $\bullet$ \textbf{Few-shot Transferability.}
To further evaluate transferability under limited supervision, we fine-tune different pre-trained models using only $10\%$ of the labeled training data, as shown in Table~\ref{tab:few_shot_results}.
Compared with CM3AE, CM3AE 2.0 improves SSIM from $63.8\%$ to $69.4\%$ on low-light enhancement, Top-1 accuracy from $32.34\%$ to $34.52\%$ on Event-only action recognition, and SR from $41.1\%$ to $43.6\%$ on Event-only tracking. Consistent gains are observed across the remaining tasks and metrics, demonstrating the effectiveness and transferability of the learned representations with limited labeled data.

\subsection{Efficiency Analysis} \label{convergeAnalysis}

Table~\ref{efficiencyMetric} compares the computational complexity and model size of different pre-training variants. From MAE to CM3AE, FLOPs increase from 287.42\,G to 391.56\,G, and the parameter count increases from 335.82\,M to 350.00\,M, mainly due to multimodal fusion and cross-modal contrastive learning. Incorporating ESMR further increases the computational cost to 404.81\,G FLOPs and the parameter count to 357.09\,M, indicating limited additional overhead for unified multimodal spatiotemporal representation learning.
The main increase comes from HCG in Stage 2. Introducing the language generation branch increases the parameter count to 8389.46\,M, while adding the diffusion branch raises FLOPs to 1682.85\,G. With both branches enabled, CM3AE 2.0 reaches 2088.18\,G FLOPs, 1044.09\,G MACs, and 9333.03\,M parameters. These costs mainly arise from the pre-trained generative models used to introduce visual generative priors and linguistic knowledge for latent cross-modal dependency learning.
Importantly, the diffusion model and language model are used only during pre-training and are discarded during downstream fine-tuning and inference. Only the pre-trained modality-specific encoders and temporal modeling components required by each downstream task are transferred to downstream networks. Therefore, the additional computational and parameter overhead of HCG does not carry over to downstream deployment.

\begin{table}[t]
\centering
\caption{Comparison of computational complexity and model size of different pre-training configurations.}
\label{efficiencyMetric}
\resizebox{\linewidth}{!}{
\begin{tabular}{l|c|c|c}
\hline
\textbf{Configuration} & \textbf{FLOPs (G)} & \textbf{MACs (G)} & \textbf{Params (M)} \\
\hline
MAE                          & 287.42  & 143.71  & 335.82  \\
CM3AE                        & 391.56  & 195.78  & 350.00  \\
\hline
CM3AE 2.0: Stage 1 only       & 404.81  & 202.41  & 357.09  \\
\quad + text generation      & 810.14  & 405.07  & 8389.46 \\
\quad + image generation     & 1682.85 & 841.43  & 1300.66 \\
\quad + both (full model)    & 2088.18 & 1044.09 & 9333.03 \\
\hline
\end{tabular}}
\end{table}

\subsection{Visualization} \label{visualization}

In this section, we visualize masked reconstruction, video-level text generation, and Event-conditioned RGB image generation during pre-training, followed by downstream predictions and attention maps. As shown in Fig.~\ref{fig:reconst}, CM3AE 2.0 reconstructs more complete object structures and clearer visual details than MAE and CM3AE, indicating improved reconstruction quality through unified multimodal representation learning. Fig.~\ref{fig:txt_generation} presents the video-level text generation results. The generated descriptions capture the main scene content and human actions, demonstrating the ability to establish connections between joint RGB-Event representations and linguistic descriptions. Fig.~\ref{fig:image_generation} further shows the RGB images generated from Event-conditioned representations. The results recover reasonable scene structures and appearance information, illustrating how visual generative priors contribute to Event representation learning.

We further visualize the downstream results in Fig.~\ref{fig:downstream_vis}, including action recognition, object detection, visual object tracking, and Event-guided low-light image enhancement. The results demonstrate the transferability of CM3AE 2.0 to diverse perception tasks under both RGB-Event and Event-only settings. Finally, Fig.~\ref{fig:attentionmaps} compares the attention maps of MAE, CM3AE, and CM3AE 2.0 across different downstream tasks. CM3AE 2.0 exhibits stronger responses in task-relevant regions and more focused attention on target objects, providing qualitative evidence of improved visual representation learning.

\subsection{Limitation Analysis} \label{limitation}

Although CM3AE 2.0 achieves consistent improvements across various RGB-Event and Event-only downstream tasks, two limitations remain. First, the framework relies on paired RGB-Event data for pre-training, while collecting high-quality synchronized multimodal data is costly. Future work could explore pre-training with unpaired or weakly paired data to leverage larger-scale data resources. Second, the hierarchical cross-modal generation strategy relies on large pre-trained diffusion and language models, introducing considerable computational overhead during pre-training. Although these models are discarded during downstream transfer and inference, developing more efficient approaches to exploit visual generative priors and linguistic knowledge for cross-modal dependency learning remains an important direction for future research.

\section{Conclusion and Future Work} \label{sec::conclusion}

In this paper, we propose CM3AE 2.0, a unified RGB Frame and Event-Voxel/-Frame pre-training framework. In the first stage, we introduce an Event-aware spatiotemporal masked reconstruction strategy to learn unified multimodal spatiotemporal representations. It models spatiotemporal relationships through masked reconstruction while exploiting cross-modal complementary information and learning shared features. In the second stage, we further propose a hierarchical cross-modal generation strategy to explore latent cross-modal dependencies. It combines event-conditioned RGB image generation and video-level text generation to leverage visual generative priors and linguistic knowledge.
We evaluate CM3AE 2.0 on four downstream tasks under both Event-only and RGB-Event settings. Extensive experiments, ablation studies, and qualitative visualizations demonstrate the effectiveness and transferability of the learned representations across diverse perception tasks, supporting the benefits of unified multimodal spatiotemporal representation learning and generative cross-modal dependency modeling.
In future work, we will explore more efficient approaches to transferring visual generative priors and linguistic knowledge while reducing pre-training overhead. We also plan to extend RGB-Event pre-training to larger-scale and more diverse datasets, including unpaired or weakly paired data, to improve the scalability and generalization of multimodal pre-training.

\small{ 
\bibliographystyle{IEEEtran}
\bibliography{reference}

\begin{thebibliography}{10}
\providecommand{\url}[1]{#1}
\csname url@samestyle\endcsname
\providecommand{\newblock}{\relax}
\providecommand{\bibinfo}[2]{#2}
\providecommand{\BIBentrySTDinterwordspacing}{\spaceskip=0pt\relax}
\providecommand{\BIBentryALTinterwordstretchfactor}{4}
\providecommand{\BIBentryALTinterwordspacing}{\spaceskip=\fontdimen2\font plus
\BIBentryALTinterwordstretchfactor\fontdimen3\font minus
  \fontdimen4\font\relax}
\providecommand{\BIBforeignlanguage}[2]{{%
\expandafter\ifx\csname l@#1\endcsname\relax
\typeout{** WARNING: IEEEtran.bst: No hyphenation pattern has been}%
\typeout{** loaded for the language `#1'. Using the pattern for}%
\typeout{** the default language instead.}%
\else
\language=\csname l@#1\endcsname
\fi
#2}}
\providecommand{\BIBdecl}{\relax}
\BIBdecl

\bibitem{gallego2020event}
G.~Gallego, T.~Delbr{\"u}ck, G.~Orchard, C.~Bartolozzi, B.~Taba, A.~Censi,
  S.~Leutenegger, A.~J. Davison, J.~Conradt, K.~Daniilidis \emph{et~al.},
  ``Event-based vision: A survey,'' \emph{IEEE transactions on pattern analysis
  and machine intelligence}, vol.~44, no.~1, pp. 154--180, 2020.

\bibitem{wang2024hardvs}
X.~Wang, Z.~Wu, B.~Jiang, Z.~Bao, L.~Zhu, G.~Li, Y.~Wang, and Y.~Tian,
  ``Hardvs: Revisiting human activity recognition with dynamic vision
  sensors,'' in \emph{Proceedings of the AAAI conference on artificial
  intelligence}, vol.~38, no.~6, 2024, pp. 5615--5623.

\bibitem{zhang2025detecting}
H.~Zhang, C.~Xu, X.~Wang, B.~Liu, G.~Hua, L.~Yu, and W.~Yang, ``Detecting every
  object from events,'' \emph{IEEE Transactions on Pattern Analysis and Machine
  Intelligence}, vol.~47, no.~8, pp. 7171--7178, 2025.

\bibitem{messikommer2025data}
N.~Messikommer, C.~Fang, M.~Gehrig, G.~Cioffi, and D.~Scaramuzza, ``Data-driven
  feature tracking for event cameras with and without frames,'' \emph{IEEE
  Transactions on Pattern Analysis and Machine Intelligence}, vol.~47, no.~5,
  pp. 3706--3717, 2025.

\bibitem{kong2024openess}
L.~Kong, Y.~Liu, L.~X. Ng, B.~R. Cottereau, and W.~T. Ooi, ``Openess:
  Event-based semantic scene understanding with open vocabularies,'' in
  \emph{2024 IEEE/CVF Conference on Computer Vision and Pattern Recognition
  (CVPR)}.\hskip 1em plus 0.5em minus 0.4em\relax IEEE, 2024, pp.
  15\,686--15\,698.

\bibitem{yang2025unsp}
J.~Yang, X.~Zhang, S.~Wang, Y.~Sun, and B.~Yin, ``Unsp: Improving
  event-to-image reconstruction with uncertainty guided self-paced learning,''
  \emph{Displays}, vol.~87, p. 102985, 2025.

\bibitem{wang2023MMPTMs}
X.~Wang, G.~Chen, G.~Qian, P.~Gao, X.-Y. Wei, Y.~Wang, Y.~Tian, and W.~Gao,
  ``Large-scale multi-modal pre-trained models: A comprehensive survey,''
  \emph{Machine Intelligence Research}, vol.~20, no.~4, pp. 447--482, 2023.

\bibitem{wang2026state}
X.~Wang, S.~Wang, Y.~Ding, Y.~Li, W.~Wu, Y.~Rong, W.~Kong, J.~Huang, S.~Li,
  H.~Yang \emph{et~al.}, ``State space model for new generation network
  alternative to transformers: a survey,'' \emph{Visual Intelligence}, vol.~4,
  no.~1, p.~29, 2026.

\bibitem{he2022mae}
K.~He, X.~Chen, S.~Xie, Y.~Li, P.~Doll{\'a}r, and R.~Girshick, ``Masked
  autoencoders are scalable vision learners,'' in \emph{Proceedings of the
  IEEE/CVF Conference on Computer Vision and Pattern Recognition}, 2022, pp.
  16\,000--16\,009.

\bibitem{zhouimage}
J.~Zhou, C.~Wei, H.~Wang, W.~Shen, C.~Xie, A.~Yuille, and T.~Kong, ``Image bert
  pre-training with online tokenizer,'' in \emph{International Conference on
  Learning Representations}, 2022.

\bibitem{oquab2023dinov2}
M.~Oquab, T.~Darcet, T.~Moutakanni, H.~Vo, M.~Szafraniec, V.~Khalidov,
  P.~Fernandez, D.~Haziza, F.~Massa, A.~El-Nouby \emph{et~al.}, ``Dinov2:
  Learning robust visual features without supervision,'' \emph{Transactions on
  Machine Learning Research}, 2024.

\bibitem{peng2022beit}
Z.~Peng, L.~Dong, H.~Bao, Q.~Ye, and F.~Wei, ``Beit v2: Masked image modeling
  with vector-quantized visual tokenizers,'' \emph{arXiv preprint
  arXiv:2208.06366}, 2022.

\bibitem{tong2022videomae}
Z.~Tong, Y.~Song, J.~Wang, and L.~Wang, ``Videomae: Masked autoencoders are
  data-efficient learners for self-supervised video pre-training,''
  \emph{Advances in neural information processing systems}, vol.~35, pp.
  10\,078--10\,093, 2022.

\bibitem{wang2023videomae}
L.~Wang, B.~Huang, Z.~Zhao, Z.~Tong, Y.~He, Y.~Wang, Y.~Wang, and Y.~Qiao,
  ``Videomae v2: Scaling video masked autoencoders with dual masking,'' in
  \emph{2023 IEEE/CVF Conference on Computer Vision and Pattern Recognition
  (CVPR)}.\hskip 1em plus 0.5em minus 0.4em\relax IEEE, 2023, pp.
  14\,549--14\,560.

\bibitem{radford2021learning}
A.~Radford, J.~W. Kim, C.~Hallacy, A.~Ramesh, G.~Goh, S.~Agarwal, G.~Sastry,
  A.~Askell, P.~Mishkin, J.~Clark \emph{et~al.}, ``Learning transferable visual
  models from natural language supervision,'' in \emph{International conference
  on machine learning}, 2021, pp. 8748--8763.

\bibitem{bachmann2022multimae}
R.~Bachmann, D.~Mizrahi, A.~Atanov, and A.~Zamir, ``Multimae: Multi-modal
  multi-task masked autoencoders,'' in \emph{European conference on computer
  vision}.\hskip 1em plus 0.5em minus 0.4em\relax Springer, 2022, pp. 348--367.

\bibitem{klenk2024masked}
S.~Klenk, D.~Bonello, L.~Koestler, N.~Araslanov, and D.~Cremers, ``Masked event
  modeling: Self-supervised pretraining for event cameras,'' in
  \emph{Proceedings of the IEEE/CVF Winter Conference on Applications of
  Computer Vision}, 2024, pp. 2378--2388.

\bibitem{yang2023event}
Y.~Yang, L.~Pan, and L.~Liu, ``Event camera data pre-training,'' in \emph{2023
  IEEE/CVF International Conference on Computer Vision (ICCV)}.\hskip 1em plus
  0.5em minus 0.4em\relax IEEE, 2023, pp. 10\,665--10\,675.

\bibitem{huang2024data}
Z.~Huang, C.~Li, H.~Chen, Y.~Deng, Y.~Geng, and L.~Wang, ``Data-efficient event
  camera pre-training via disentangled masked modeling,'' \emph{arXiv preprint
  arXiv:2403.00416}, 2024.

\bibitem{mohammadi2025tespec}
M.~Mohammadi, Z.~Wu, and I.~Gilitschenski, ``Tespec: Temporally-enhanced
  self-supervised pretraining for event cameras,'' in \emph{2025 IEEE/CVF
  International Conference on Computer Vision (ICCV)}.\hskip 1em plus 0.5em
  minus 0.4em\relax IEEE, 2025, pp. 7782--7793.

\bibitem{zhou2024eventbind}
J.~Zhou, X.~Zheng, Y.~Lyu, and L.~Wang, ``Eventbind: Learning a unified
  representation to bind them all for event-based open-world understanding,''
  in \emph{European Conference on Computer Vision}, 2024, pp. 477--494.

\bibitem{jeong2026cross}
S.~Jeong, H.~Chen, S.~Yun, S.~Cho, W.~Huang, X.~Liu, and M.~Imani,
  ``Cross-modal event encoder: Bridging image--text knowledge to event
  streams,'' in \emph{2026 IEEE/CVF Winter Conference on Applications of
  Computer Vision (WACV)}.\hskip 1em plus 0.5em minus 0.4em\relax IEEE, 2026,
  pp. 3213--3222.

\bibitem{chen2026scaling}
Z.~Chen, J.~Hou, Z.~Zhu, J.~Wu, and G.~Shi, ``Scaling dense event-stream
  pretraining from visual foundation models,'' in \emph{Proceedings of the
  IEEE/CVF Conference on Computer Vision and Pattern Recognition}.\hskip 1em
  plus 0.5em minus 0.4em\relax IEEE, 2026, pp. 8011--8022.

\bibitem{liu2025eventgpt}
S.~Liu, J.~Li, G.~Zhao, Y.~Zhang, X.~Meng, F.~R. Yu, X.~Ji, and M.~Li,
  ``Eventgpt: Event stream understanding with multimodal large language
  models,'' in \emph{2025 IEEE/CVF Conference on Computer Vision and Pattern
  Recognition (CVPR)}.\hskip 1em plus 0.5em minus 0.4em\relax IEEE, 2025, pp.
  29\,139--29\,149.

\bibitem{wu2025cm3ae}
W.~Wu, X.~Wang, C.~Li, B.~Jiang, J.~Tang, B.~Luo, and Q.~Liu, ``Cm3ae: A
  unified rgb frame and event-voxel/-frame pre-training framework,'' in
  \emph{Proceedings of the 33rd ACM International Conference on Multimedia},
  2025, pp. 2159--2168.

\bibitem{ren2025rethinking}
H.~Ren, Y.~Zhou, J.~Zhu, X.~Lin, H.~Fu, Y.~Huang, Y.~Fang, F.~Ma, H.~Yu, and
  B.~Cheng, ``Rethinking efficient and effective point-based networks for event
  camera classification and regression,'' \emph{IEEE Transactions on Pattern
  Analysis and Machine Intelligence}, 2025.

\bibitem{cao2025exploiting}
M.~Cao, X.~Shu, X.~Jiang, R.~Yan, Y.~Yao, and J.~Tang, ``Exploiting frequency
  dynamics for enhanced multimodal event-based action recognition,'' in
  \emph{2025 IEEE/CVF International Conference on Computer Vision
  (ICCV)}.\hskip 1em plus 0.5em minus 0.4em\relax IEEE, 2025, pp. 5969--5979.

\bibitem{wang2026event}
X.~Wang, S.~Wang, P.~Shao, L.~Zhu, B.~Jiang, and Y.~Tian, ``Event stream based
  human action recognition: A high-definition benchmark dataset and
  algorithms,'' \emph{International Journal of Computer Vision}, vol. 134,
  no.~4, p. 181, 2026.

\bibitem{gehrig2023recurrent}
M.~Gehrig and D.~Scaramuzza, ``Recurrent vision transformers for object
  detection with event cameras,'' in \emph{2023 IEEE/CVF Conference on Computer
  Vision and Pattern Recognition (CVPR)}.\hskip 1em plus 0.5em minus
  0.4em\relax IEEE, 2023, pp. 13\,884--13\,893.

\bibitem{zhu2026rethinking}
L.~Zhu, T.~Long, X.~Wang, L.~Wang, and H.~Huang, ``Rethinking scale-aware
  temporal encoding for event-based object detection,'' \emph{Advances in
  Neural Information Processing Systems}, vol.~38, pp. 144\,008--144\,032,
  2026.

\bibitem{wang2025object}
X.~Wang, Y.~Jin, W.~Wu, W.~Zhang, L.~Zhu, B.~Jiang, and Y.~Tian, ``Object
  detection using event camera: A moe heat conduction based detector and a new
  benchmark dataset,'' in \emph{2025 IEEE/CVF Conference on Computer Vision and
  Pattern Recognition (CVPR)}.\hskip 1em plus 0.5em minus 0.4em\relax IEEE,
  2025, pp. 29\,321--29\,330.

\bibitem{wang2024event}
X.~Wang, S.~Wang, C.~Tang, L.~Zhu, B.~Jiang, Y.~Tian, and J.~Tang, ``Event
  stream-based visual object tracking: A high-resolution benchmark dataset and
  a novel baseline,'' in \emph{2024 IEEE/CVF Conference on Computer Vision and
  Pattern Recognition (CVPR)}.\hskip 1em plus 0.5em minus 0.4em\relax IEEE,
  2024, pp. 19\,248--19\,257.

\bibitem{zhu2024continuous}
L.~Zhu, X.~Chen, L.~Wang, X.~Wang, Y.~Tian, and H.~Huang, ``Continuous-time
  object segmentation using high temporal resolution event camera,'' \emph{IEEE
  Transactions on Pattern Analysis and Machine Intelligence}, vol.~47, no.~2,
  pp. 807--824, 2024.

\bibitem{liang2024towards}
G.~Liang, K.~Chen, H.~Li, Y.~Lu, and L.~Wang, ``Towards robust event-guided
  low-light image enhancement: A large-scale real-world event-image dataset and
  novel approach,'' in \emph{2024 IEEE/CVF Conference on Computer Vision and
  Pattern Recognition (CVPR)}.\hskip 1em plus 0.5em minus 0.4em\relax IEEE,
  2024, pp. 23--33.

\bibitem{weng2024event}
J.~Weng, B.~Li, and K.~Huang, ``Event-based image enhancement under high
  dynamic range scenarios,'' in \emph{Asian Conference on Computer
  Vision}.\hskip 1em plus 0.5em minus 0.4em\relax Springer, 2024, pp. 346--360.

\bibitem{sun2025low}
L.~Sun, Y.~Bao, J.~Zhai, J.~Liang, Y.~Zhang, K.~Wang, D.~P. Paudel, and
  L.~Van~Gool, ``Low-light image enhancement using event-based illumination
  estimation,'' in \emph{2025 IEEE/CVF International Conference on Computer
  Vision (ICCV)}.\hskip 1em plus 0.5em minus 0.4em\relax IEEE, 2025, pp.
  6667--6677.

\bibitem{sun2025exploring}
C.~Sun, J.~Zhang, Y.~Wang, H.~Ge, Q.~Xia, B.~Yin, and X.~Yang, ``Exploring
  historical information for rgbe visual tracking with mamba,'' in \emph{2025
  IEEE/CVF Conference on Computer Vision and Pattern Recognition (CVPR)}.\hskip
  1em plus 0.5em minus 0.4em\relax IEEE, 2025, pp. 6500--6509.

\bibitem{sun2026aligntrack}
C.~Sun, J.~Zhang, Y.~Wang, Y.~Wang, Y.~Jiang, B.~Yin, and X.~Yang,
  ``Aligntrack: Top-down spatiotemporal resolution alignment for rgb-event
  visual tracking,'' in \emph{Proceedings of the AAAI Conference on Artificial
  Intelligence}, vol.~40, no.~11, 2026, pp. 9171--9179.

\bibitem{zhu2025crsot}
Y.~Zhu, X.~Wang, C.~Li, B.~Jiang, L.~Zhu, Z.~Huang, Y.~Tian, and J.~Tang,
  ``Crsot: Cross-resolution object tracking using unaligned frame and event
  cameras,'' \emph{IEEE Transactions on Multimedia}, 2025.

\bibitem{lu2026flexevent}
D.~Lu, L.~Kong, G.~H. Lee, C.~Simon~Chane, and W.~Ooi, ``Flexevent: towards
  flexible event-frame object detection at varying operational frequencies,''
  \emph{Advances in Neural Information Processing Systems}, vol.~38, pp.
  89\,009--89\,044, 2026.

\bibitem{wang2026beyond}
K.~Wang, S.~Liu, H.~Shi, L.~Shi, and H.~Chen, ``Beyond duality: A hybrid
  framework of leveraging shared and private features for rgb-event object
  detection,'' in \emph{Proceedings of the IEEE/CVF Conference on Computer
  Vision and Pattern Recognition}, 2026, pp. 4415--4424.

\bibitem{wang2026human}
S.~Wang, X.~Wang, B.~Jiang, L.~Zhu, G.~Li, Y.~Wang, Y.~Tian, and J.~Tang,
  ``Human activity recognition using rgb-dvs cameras: A multi-modal heat
  conduction model and a benchmark dataset,'' \emph{International Journal of
  Computer Vision}, vol. 134, no.~8, p. 377, 2026.

\bibitem{wang2026rgb}
X.~Wang, H.~Wang, S.~Wang, Q.~Chen, J.~Jin, H.~Song, B.~Jiang, and C.~Li,
  ``Rgb-event based pedestrian attribute recognition: A benchmark dataset and
  an asymmetric rwkv fusion framework,'' in \emph{Proceedings of the IEEE/CVF
  Conference on Computer Vision and Pattern Recognition}, 2026, pp.
  32\,745--32\,755.

\bibitem{wang2026injecting}
H.~Wang, H.~Zhou, H.~Liu, and L.~Yan, ``Injecting frame-event complementary
  fusion into diffusion for optical flow in challenging scenes,''
  \emph{Advances in Neural Information Processing Systems}, vol.~38, pp.
  31\,614--31\,633, 2026.

\bibitem{he2020momentum}
K.~He, H.~Fan, Y.~Wu, S.~Xie, and R.~Girshick, ``Momentum contrast for
  unsupervised visual representation learning,'' in \emph{2020 IEEE/CVF
  conference on computer vision and pattern recognition (CVPR)}.\hskip 1em plus
  0.5em minus 0.4em\relax IEEE, 2020, pp. 9726--9735.

\bibitem{wu2026vehicle}
W.~Wu, X.~Wang, C.~Li, J.~Tang, and B.~Luo, ``Vehicle-centric perception via
  multimodal structured pre-training,'' \emph{IEEE Transactions on Circuits and
  Systems for Video Technology}, vol.~36, pp. 8615--8630, 2026.

\bibitem{fini2025multimodal}
E.~Fini, M.~Shukor, X.~Li, P.~Dufter, M.~Klein, D.~Haldimann, S.~Aitharaju,
  V.~G.~T. da~Costa, L.~B{\'e}thune, Z.~Gan \emph{et~al.}, ``Multimodal
  autoregressive pre-training of large vision encoders,'' in \emph{2025
  IEEE/CVF Conference on Computer Vision and Pattern Recognition (CVPR)}.\hskip
  1em plus 0.5em minus 0.4em\relax IEEE, 2025, pp. 9641--9654.

\bibitem{stracke2025cleandift}
N.~Stracke, S.~A. Baumann, K.~Bauer, F.~Fundel, and B.~Ommer, ``Cleandift:
  Diffusion features without noise,'' in \emph{2025 IEEE/CVF Conference on
  Computer Vision and Pattern Recognition (CVPR)}.\hskip 1em plus 0.5em minus
  0.4em\relax IEEE, 2025, pp. 117--127.

\bibitem{zhang2026vision}
T.~Zhang, Y.~Li, Y.-C. Chou, J.~Chen, A.~Yuille, C.~Wei, and J.~Xiao,
  ``Vision-language-vision auto-encoder: Scalable knowledge distillation from
  diffusion models,'' \emph{Advances in Neural Information Processing Systems},
  vol.~38, pp. 79\,174--79\,202, 2026.

\bibitem{liang2025efficient}
Q.~Liang, Q.~Li, S.~Liu, X.~Cao, J.~Lu, F.~Yang, W.~Zhang, K.~Huang, and
  Y.~Tian, ``Efficient event camera data pretraining with adaptive prompt
  fusion,'' in \emph{2025 IEEE/CVF International Conference on Computer Vision
  (ICCV)}.\hskip 1em plus 0.5em minus 0.4em\relax IEEE, 2025, pp. 8656--8667.

\bibitem{lu2026eventdrive}
D.~Lu, R.~Li, A.~Liang, L.~Kong, W.~Yin, L.~X. Ng, B.~R. Cottereau, C.~S.
  Chane, and W.~T. Ooi, ``Eventdrive: Event cameras for vision-language driving
  intelligence,'' in \emph{Proceedings of the IEEE/CVF Conference on Computer
  Vision and Pattern Recognition}, 2026, pp. 22\,312--22\,322.

\bibitem{rombach2022high}
R.~Rombach, A.~Blattmann, D.~Lorenz, P.~Esser, and B.~Ommer, ``High-resolution
  image synthesis with latent diffusion models,'' in \emph{Proceedings of the
  IEEE/CVF Conference on Computer Vision and Pattern Recognition (CVPR)}, 2022,
  pp. 10\,684--10\,695.

\bibitem{bai2025qwen25vl}
S.~Bai, K.~Chen, X.~Liu, J.~Wang, W.~Ge, S.~Song, K.~Dang, P.~Wang, S.~Wang,
  J.~Tang, H.~Zhong, Y.~Zhu, M.~Yang, Z.~Li, J.~Wan, P.~Wang, W.~Ding, Z.~Fu,
  Y.~Xu, J.~Ye, X.~Zhang, T.~Xie, Z.~Cheng, H.~Zhang, Z.~Yang, H.~Xu, and
  J.~Lin, ``Qwen2.5-vl technical report,'' \emph{arXiv preprint
  arXiv:2502.13923}, 2025.

\bibitem{grattafiori2024llama}
A.~Grattafiori, A.~Dubey, A.~Jauhri, A.~Pandey, A.~Kadian, A.~Al-Dahle,
  A.~Letman, A.~Mathur, A.~Schelten, A.~Vaughan \emph{et~al.}, ``The llama 3
  herd of models,'' \emph{arXiv preprint arXiv:2407.21783}, 2024.

\bibitem{bertasius2021space}
G.~Bertasius, H.~Wang, and L.~Torresani, ``Is space-time attention all you need
  for video understanding?'' \emph{arXiv preprint arXiv:2102.05095}, 2021.

\bibitem{li2022exploring}
Y.~Li, H.~Mao, R.~Girshick, and K.~He, ``Exploring plain vision transformer
  backbones for object detection,'' in \emph{European conference on computer
  vision}.\hskip 1em plus 0.5em minus 0.4em\relax Springer, 2022, pp. 280--296.

\bibitem{zheng2024odtrack}
Y.~Zheng, B.~Zhong, Q.~Liang, Z.~Mo, S.~Zhang, and X.~Li, ``Odtrack: Online
  dense temporal token learning for visual tracking,'' in \emph{Proceedings of
  the AAAI conference on artificial intelligence}, vol.~38, no.~7, 2024, pp.
  7588--7596.

\bibitem{tang2025revisiting}
C.~Tang, X.~Wang, J.~Huang, B.~Jiang, L.~Zhu, S.~Chen, J.~Zhang, Y.~Wang, and
  Y.~Tian, ``Revisiting color-event based tracking: A unified network, dataset,
  and metric,'' \emph{Pattern Recognition}, p. 112718, 2025.

\bibitem{gehrig2024low}
D.~Gehrig and D.~Scaramuzza, ``Low-latency automotive vision with event
  cameras,'' \emph{Nature}, vol. 629, no. 8014, pp. 1034--1040, 2024.

\bibitem{fang2023eva}
Y.~Fang, W.~Wang, B.~Xie, Q.~Sun, L.~Wu, X.~Wang, T.~Huang, X.~Wang, and
  Y.~Cao, ``Eva: Exploring the limits of masked visual representation learning
  at scale,'' in \emph{2023 IEEE/CVF conference on computer vision and pattern
  recognition (CVPR)}.\hskip 1em plus 0.5em minus 0.4em\relax IEEE, 2023, pp.
  19\,358--19\,369.

\bibitem{liupixmim}
Y.~Liu, S.~Zhang, J.~Chen, K.~Chen, and D.~Lin, ``Pixmim: Rethinking pixel
  reconstruction in masked image modeling,'' \emph{Transactions on Machine
  Learning Research}.

\bibitem{yang2024event}
Y.~Yang, L.~Pan, and L.~Liu, ``Event camera data dense pre-training,'' in
  \emph{European Conference on Computer Vision}.\hskip 1em plus 0.5em minus
  0.4em\relax Springer, 2024, pp. 292--310.

\bibitem{kim2021nimagenet}
J.~Kim, J.~Bae, G.~Park, D.~Zhang, and Y.~M. Kim, ``N-imagenet: Towards robust,
  fine-grained object recognition with event cameras,'' in \emph{Proceedings of
  the IEEE/CVF international conference on computer vision}, 2021, pp.
  2146--2156.

\bibitem{wang2023visevent}
X.~Wang, J.~Li, L.~Zhu, Z.~Zhang, Z.~Chen, X.~Li, Y.~Wang, Y.~Tian, and F.~Wu,
  ``Visevent: Reliable object tracking via collaboration of frame and event
  flows,'' \emph{IEEE Transactions on Cybernetics}, vol.~54, no.~3, pp.
  1997--2010, 2023.

\bibitem{paszke2019pytorch}
A.~Paszke, S.~Gross, F.~Massa, A.~Lerer, J.~Bradbury, G.~Chanan, T.~Killeen,
  Z.~Lin, N.~Gimelshein, L.~Antiga \emph{et~al.}, ``Pytorch: An imperative
  style, high-performance deep learning library,'' \emph{Advances in neural
  information processing systems}, vol.~32, 2019.

\bibitem{adamW}
I.~Loshchilov and F.~Hutter, ``Decoupled weight decay regularization,'' in
  \emph{International Conference on Learning Representations}, 2019.

\end{thebibliography}
}

\vspace{-0.5cm}
\begin{IEEEbiography} [{\includegraphics[width=1in,height=1.25in,clip,keepaspectratio]{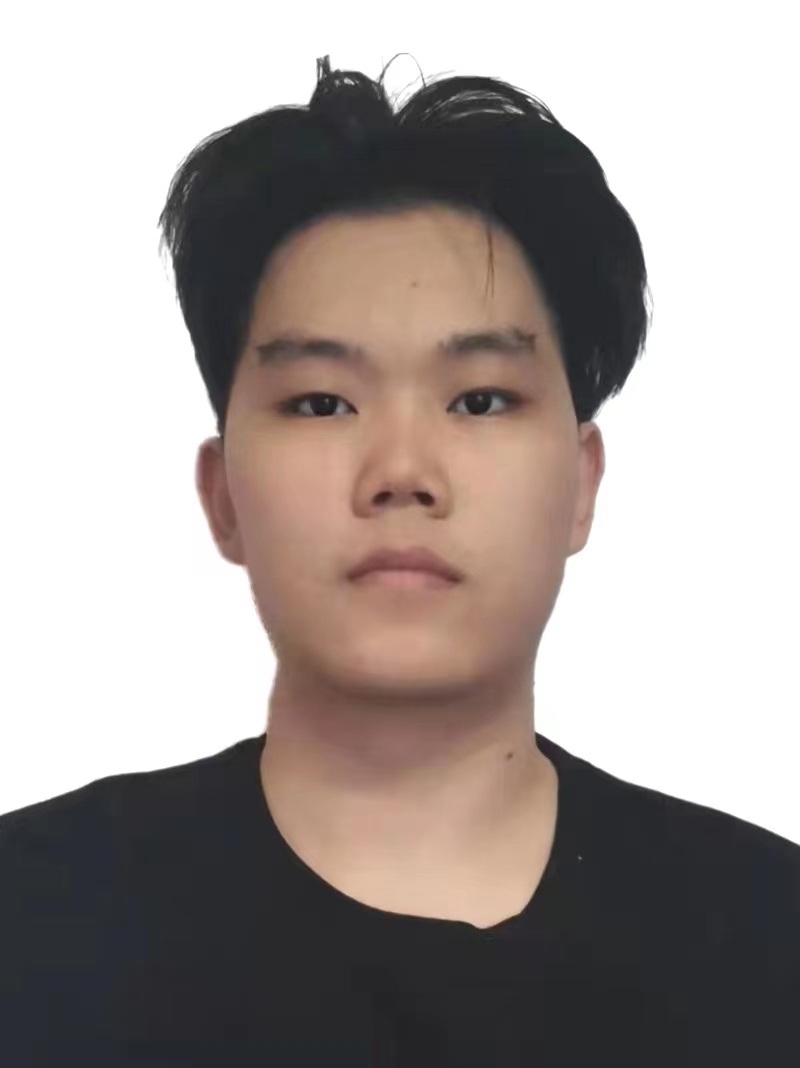}}]
{Wentao Wu} received the B.S. degree from Jiangxi University of Science and Technology, Nanchang, China, in 2022. He is currently working toward the Ph.D. degree with the school of Artificial Intelligence, Anhui University. His research interests include computer vision and deep learning.
\end{IEEEbiography} 

\vspace{-0.5cm}
\begin{IEEEbiography} [{\includegraphics[width=1in,height=1.25in,clip,keepaspectratio]{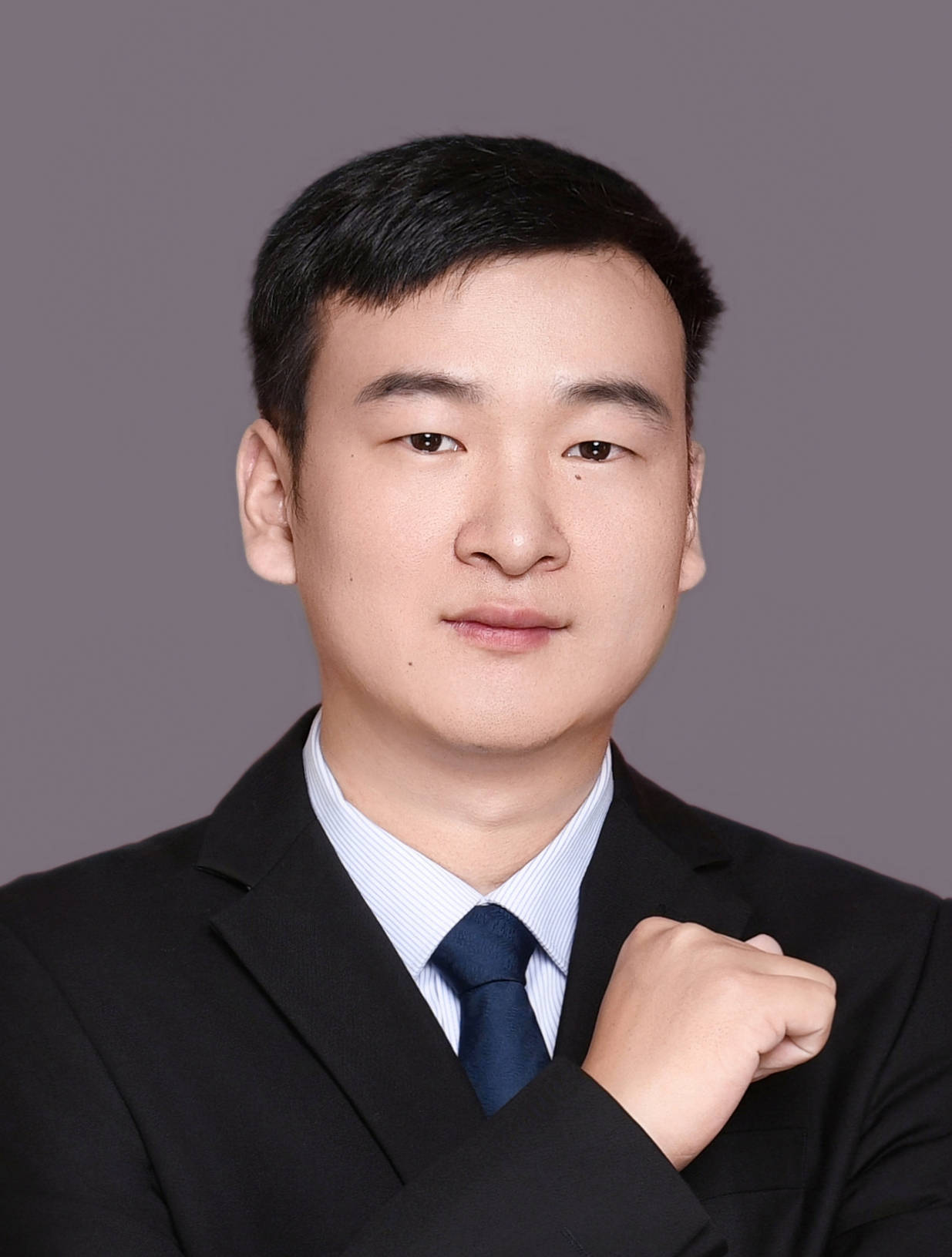}}]
{Xiao Wang} (Member, IEEE) received the B.S. degree from West Anhui University, Luan, China, in 2013. He received the Ph.D. degree in computer science from Anhui University, Hefei, China, in 2019. From 2015 and 2016, he was a visiting student at the School of Data and Computer Science, Sun Yat-sen University, Guangzhou, China. He also has a visit to UBTECH Sydney Artificial Intelligence Centre, the Faculty of Engineering, the University of Sydney, in 2019. He finished the postdoc research at Peng Cheng Laboratory, Shenzhen, China, from April 2020 to April 2022. He is now an Associate Professor at the School of Computer Science and Technology, Anhui University, Hefei, China. His current research interests are mainly Computer Vision, Event-based Vision, Machine Learning, and Pattern Recognition. He serves as a reviewer for a number of journals and conferences, such as IEEE TCSVT, TNNLS, TIP, IJCV, CVIU, PR, CVPR, ICCV, AAAI, ECCV, ACCV, ACM-MM, and WACV. \textbf{Homepage}: \url{https://wangxiao5791509.github.io/} 
\end{IEEEbiography} 

\vspace{-0.5cm}
\begin{IEEEbiography} [{\includegraphics[width=1in,height=1.25in,clip,keepaspectratio]{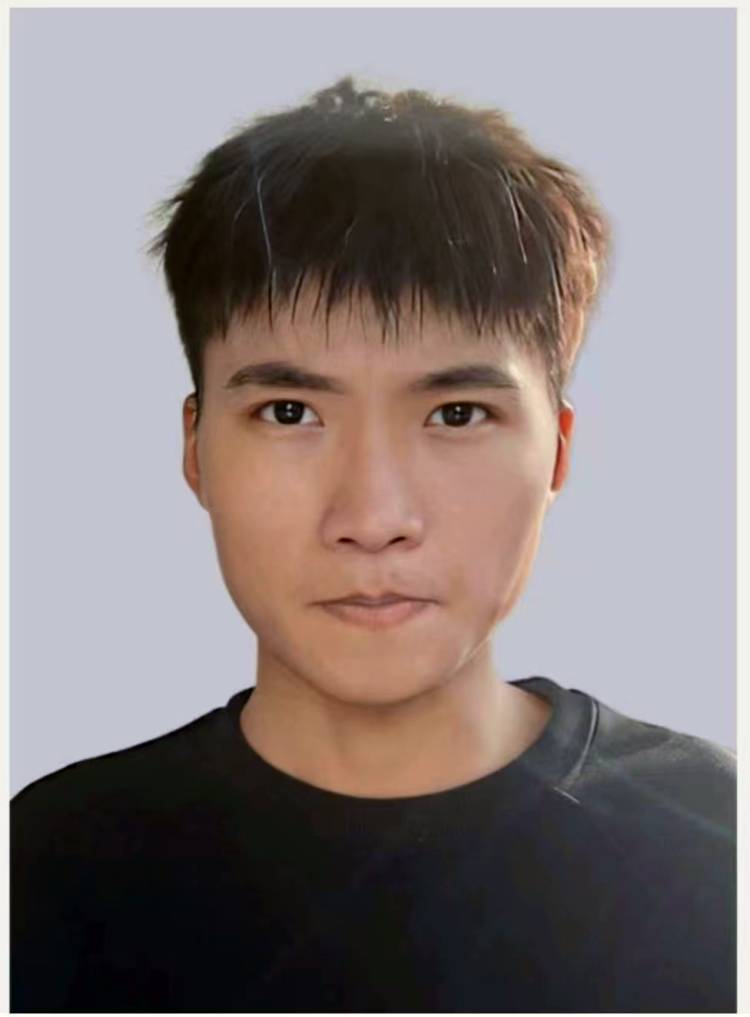}}]
{Yi Wang} received the B.S. degree from Anhui Normal University, Wuhu, China. He is currently working toward the M.S. degree with the School of Computer Science and Technology, Anhui University, Hefei, China. His research interests include time-series classification, signal processing, and edge-localized mode (ELM) recognition.  
\end{IEEEbiography}

\vspace{-0.5cm}
\begin{IEEEbiography} [{\includegraphics[width=1in,height=1.25in,clip,keepaspectratio]{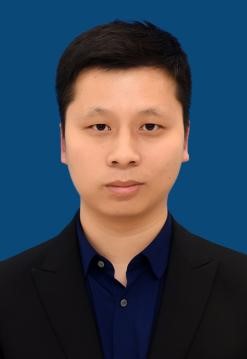}}]
{Chenglong Li} (Senior Member, IEEE) received the M.S. and Ph.D. degrees from the School of Computer Science and Technology,Anhui University, Hefei, China, in 2013 and 2016, respectively. From 2014 to 2015, he was a Visiting Student at the School of Artificial Intelligence, Sun Yat-sen University, Guangzhou, China. He was a Post-Doctoral Research Fellow with the Center for Research on Intelligent Perception and Computing (CRIPAC), National Laboratory of Pattern Recognition (NLPR), Institute of Automation, Chinese Academy of Sciences (CASIA), Beijing, China. He is currently a Professor and a Ph.D. Supervisor with the School of Artificial Intelligence, Anhui University. His research interests include computer vision and deep learning
\end{IEEEbiography}

\vspace{-0.5cm}
\begin{IEEEbiography}[{\includegraphics[width=1in,height=1.25in,clip,keepaspectratio]{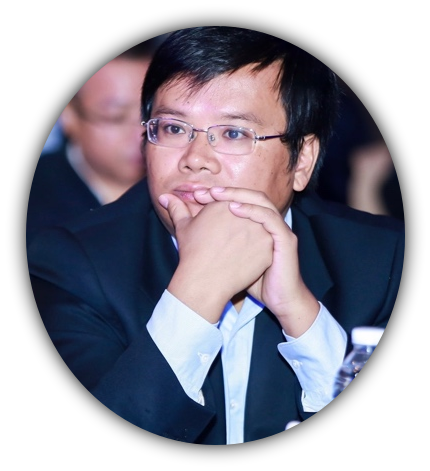}}] 
{Yonghong Tian} (Fellow, IEEE) is currently a Boya Distinguished Professor with the Department of Computer Science and Technology, Peking University, China, and is also the deputy director of Artificial Intelligence Research Center, PengCheng Laboratory, Shenzhen, China. His research interests include neuromorphic vision, brain-inspired computation and multimedia big data. He is the author or coauthor of over 200 technical articles in refereed journals such as IEEE TPAMI/TNNLS/TIP/TMM/TCSVT/TKDE/TPDS, ACM CSUR/TOIS/TOMM and conferences such as NeurIPS/CVPR/ICCV/AAAI/ACMMM/WWW. Prof. Tian was/is an Associate Editor of IEEE TCSVT (2018.1-), IEEE TMM (2014.8-2018.8), IEEE Multimedia Mag. (2018.1-), and IEEE Access (2017.1-). He co-initiated IEEE Int’l Conf. on Multimedia Big Data (BigMM) and served as the TPC Co-chair of BigMM 2015, and aslo served as the Technical Program Co-chair of IEEE ICME 2015, IEEE ISM 2015 and IEEE MIPR 2018/2019, and General Co-chair of IEEE MIPR 2020 and ICME2021. He is the steering member of IEEE ICME (2018-) and IEEE BigMM (2015-), and is a TPC Member of more than ten conferences such as CVPR, ICCV, ACM KDD, AAAI, ACM MM and ECCV. He was the recipient of the Chinese National Science Foundation for Distinguished Young Scholars in 2018, two National Science and Technology Awards and three ministerial-level awards in China, and obtained the 2015 EURASIP Best Paper Award for Journal on Image and Video Processing, and the best paper award of IEEE BigMM 2018. He is a senior member of IEEE, CIE and CCF, a member of ACM.
\end{IEEEbiography}

\vspace{-0.5cm}
\begin{IEEEbiography} [{\includegraphics[width=1in,height=1.25in,clip,keepaspectratio]{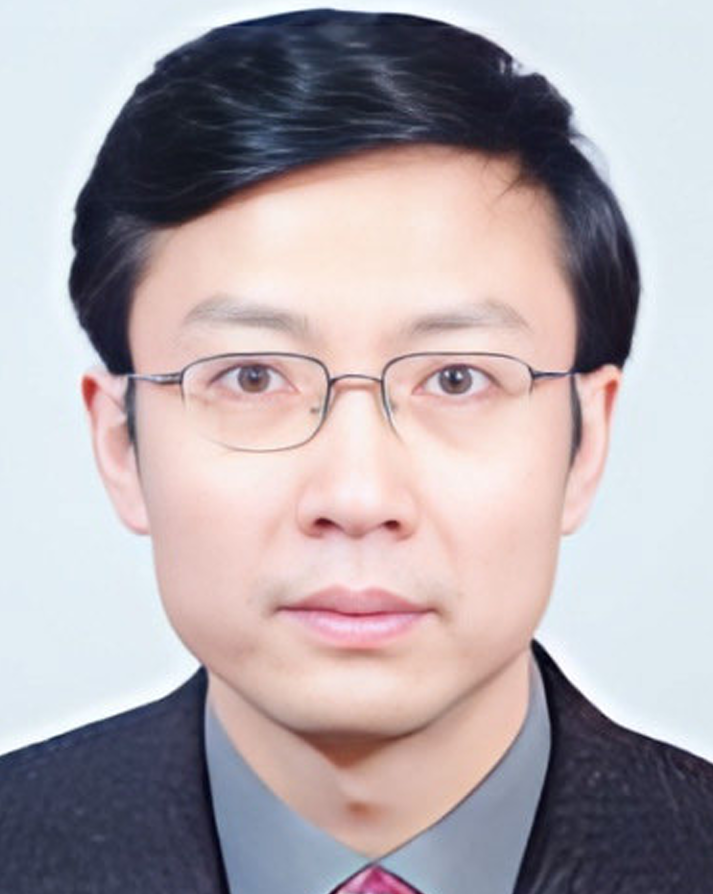}}] 
{Bin Luo} (Senior Member, IEEE) received the B.Eng. degree in electronics and the M.Eng. degree in computer science from Anhui University, Hefei, China, in 1984 and 1991, respectively, and the Ph.D. degree in computer science from the University of York, York, U.K., in 2002. From 2000 to 2004, he was a Research Associate with the University of York. He is currently a Professor with Anhui University. His research interests include graph spectral analysis, large image database retrieval, image and graph matching, statistical pattern recognition, digital watermarking, and information security.
\end{IEEEbiography}

\end{CJK}
\end{document}